\documentclass[10pt]{article} 
\usepackage[preprint]{tmlr}

\usepackage{amsmath,amsfonts,bm}

\def\eqref#1{equation~\ref{#1}}

\def\1{\bm{1}}

\DeclareMathAlphabet{\mathsfit}{\encodingdefault}{\sfdefault}{m}{sl}
\SetMathAlphabet{\mathsfit}{bold}{\encodingdefault}{\sfdefault}{bx}{n}

\usepackage{hyperref}
\usepackage{url}

\usepackage{float}
\usepackage{graphicx}
\usepackage{enumitem}
\usepackage{booktabs}
 \usepackage{svg}
\usepackage{amsthm}
\newtheorem{definition}{Definition}[section]
\newtheorem{assumption}{Assumption}[section]
\newtheorem{theorem}{Theorem}[section]

\newtheorem{example}[theorem]{Example}

\title{A Unified Theory of Sparse Dictionary Learning in Mechanistic Interpretability: Piecewise Biconvexity and Spurious Minima}

\author{\name Yiming Tang \email yiming@nus.edu.sg \\
      National University of Singapore
      \AND
      \name Harshvardhan Saini  \email \\
      Indian Institute of Technology, Dhanbad
      \AND
      \name Zhaoqian Yao  \email  \\
      Chinese University of Hong Kong
      \AND
      \name Zheng Lin \email  \\
      Hong Kong University of Science and Technology
      \AND
      \name Yizhen Liao  \email \\
      National University of Singapore
      \AND
      \name Jingyi Cui  \email \\
      Peking University
      \AND
      \name Yisen Wang \email  \\
      Peking University
      \AND
      \name Mengnan Du \email  \\
      Chinese University of Hong Kong
      \AND
      \name Dianbo Liu \email dianbo@nus.edu.sg \\
      National University of Singapore
      }

\def\month{MM}  
\def\year{YYYY} 
\def\openreview{\url{https://openreview.net/forum?id=XXXX}} 

\begin{document}

\maketitle

\begin{abstract}
As AI models achieve remarkable capabilities across diverse domains, understanding what representations they learn and how they encode concepts has become increasingly important for both scientific progress and trustworthy deployment. Recent works in mechanistic interpretability have widely reported that neural networks represent meaningful concepts as linear directions in their representation spaces and often encode diverse concepts in superposition. Various sparse dictionary learning (SDL) methods, including sparse autoencoders, transcoders, and crosscoders, are utilized to address this by training auxiliary models with sparsity constraints to disentangle these superposed concepts into monosemantic features. These methods are the backbone of modern mechanistic interpretability, yet in practice they consistently produce polysemantic features, feature absorption, and dead neurons, with very limited theoretical understanding of why these phenomena occur. Existing theoretical work is limited to tied-weight sparse autoencoders, leaving the broader family of SDL methods without formal grounding. We develop the first unified theoretical framework that casts all major SDL variants as a single piecewise biconvex optimization problem, and characterize its global solution set, non-identifiability, and spurious optima. This analysis yields principled explanations for feature absorption and dead neurons. To expose these pathologies under full ground-truth access, we introduce the Linear Representation Bench. Guided by our theory, we propose feature anchoring, a novel technique that restores SDL identifiability, substantially improving feature recovery across synthetic benchmarks and real neural representations.
\end{abstract}
\section{Introduction}
\label{sec:introduction}

As artificial intelligence systems scale to frontier capabilities, understanding their internal mechanisms has become essential for safe deployment \citep{lipton2017mythosmodelinterpretability, rudin2019stopexplainingblackbox}. A central insight from mechanistic interpretability is that neural networks encode interpretable concepts as linear directions in superposition \citep{park2024linearrepresentationhypothesisgeometry, elhage2022toymodelssuperposition}, where individual neurons respond to multiple unrelated concepts—a phenomenon known as polysemanticity \citep{bricken2023monosemanticity, templeton2024scaling}. To disentangle these superposed representations, Sparse Dictionary Learning methods, including sparse autoencoders (SAEs) \citep{cunningham2023sparseautoencodershighlyinterpretable}, transcoders \citep{dunefsky2024transcodersinterpretablellmfeature}, and crosscoders \citep{lindsey2024crosscoders}, have emerged as the dominant paradigm, achieving remarkable empirical success on frontier language models \citep{templeton2024scaling, gao2024scalingevaluatingsparseautoencoders} and enabling applications from feature steering \citep{wang2025personafeaturescontrolemergent} to circuit analysis \citep{marks2025sparsefeaturecircuitsdiscovering} and medical diagnosis \citep{abdulaal2024xrayworth15features}.

Despite this empirical success across diverse applications, SDL methods consistently exhibit persistent failure modes: learned features remain polysemantic \citep{chanin2025absorptionstudyingfeaturesplitting}, "dead neurons" fail to activate on any data samples \citep{bricken2023monosemanticity}, and "feature absorption" occurs where one neuron captures specific sub-concepts while another responds to the remaining related concepts \citep{chanin2025absorptionstudyingfeaturesplitting}. Practitioners have developed techniques to address these issues, including neuron resampling \citep{bricken2023monosemanticity}, auxiliary losses \citep{gao2024scalingevaluatingsparseautoencoders}, and careful hyperparameter tuning, yet these fixes remain ad-hoc engineering solutions without principled justification. Critically, these phenomena persist even with careful training on clean data, suggesting they are not mere implementation artifacts but reflect fundamental structural properties of the SDL optimization problem itself.

We argue that the root cause of these failure modes is the non-identifiability of SDL methods: even under the idealized Linear Representation Hypothesis, SDL optimization admits multiple solutions achieving perfect reconstruction loss, some recovering no interpretable ground-truth features at all, necessitating comprehensive theoretical analysis on SDL methods. While classical dictionary learning theory provides identifiability guarantees under strict conditions \citep{spielman2012exactrecoverysparselyuseddictionaries, gribonval2010dictionaryidentificationsparse}, and recent work establishes necessary conditions for tied-weight SAEs \citep{cui2025theoreticalunderstandingidentifiablesparse}, no unified theoretical framework explains why diverse SDL methods, SAEs, transcoders, crosscoders, and their variants, systematically fail in predictable ways. Without theoretical grounding, the development of improved SDL methods remains largely empirical, potentially leaving highly effective techniques underexplored.

In this work, we develop a unified theoretical framework that formalizes SDL as a general optimization problem, encompassing various SDL methods \citep{bussmann2024batchtopksparseautoencoders,bussmann2025learningmultilevelfeaturesmatryoshka,tang2025doesmodelfailautomatic,luo2023promptengineeringlensoptimal} as special cases. We demonstrate how these diverse methods instantiate our framework through different choices of input-output representation pairs, activation functions, and loss designs. We establish rigorous conditions under which SDL methods provably recover ground-truth interpretable features, characterizing the roles of feature sparsity, latent dimensionality, and activation functions. Through detailed analysis of the optimization landscape, we demonstrate that global minima correspond to correct feature recovery and provide necessary and sufficient conditions for achieving zero loss. We establish the prevalence of spurious partial minima exhibiting un-disentangled polysemanticity, providing novel theoretical explanations for feature absorption \citep{chanin2025absorptionstudyingfeaturesplitting} and the effectiveness of neuron resampling \citep{bricken2023monosemanticity}. We design the Linear Representation Bench, a synthetic benchmark that strictly follows the Linear Representation Hypothesis, to evaluate SDL methods with fully accessible ground-truth features. Motivated by our theoretical insights, we propose feature anchoring, a technique applicable to all SDL methods which achieves improved feature recovery by constraining learned features to known anchor directions.

Our main contributions are as follows:

\begin{itemize}
    \item We build the first theoretical framework for SDL in mechanistic interpretability as a general optimization problem encompassing diverse SDL methods.

    \item We theoretically prove that SDL optimization is biconvex, bridging mechanistic interpretability methods with traditional biconvex optimization theory.

    \item We characterize SDL optimization landscape and prove its non-identifiability, providing novel explanations for various phenomena observed empirically.
    
    \item We design the Linear Representation Bench, a benchmark with fully accessible ground-truth features, enabling fully transparent evaluation of SDL methods.
    
    \item We propose a novel technique, feature anchoring, that can achieve largely improved feature recovery performance applicable for all SDL methods. We validate the effectiveness of feature anchoring with extensive experiments across diverse SDL methods and settings.
    
\end{itemize}

\section{Preliminaries}
\label{sec:preliminaries}

In this section, we present a unified theoretical framework for Sparse Dictionary Learning (SDL). We begin with the formal definitions of foundational concepts and our assumptions on representations. Then we introduce our framework and how various SDL methods instantiate it.

\subsection{Input Distribution and Model Representation}

\begin{definition}[Input Distribution]
\label{def:input_distribution}
Let $\mathcal{D}$ denote the distribution over possible inputs, $\mathcal{X}$, to a neural network. For example, $\mathcal{D}$ could be the distribution of natural images or the distribution of text sequences (Notations in Appendix~\ref{app:notation}).
\end{definition}

\begin{definition}[Model Representation]
\label{def:representation}
For a given model representation $\mathbf{x}$, let $n \in \mathbb{N}$ be its dimensionality. For each $s \sim \mathcal{D}$, the network produces a representation vector $\mathbf{x}(s)$, which is directly observable by running the model on $s$.
\end{definition}

\subsection{Assumptions}
Empirical studies in mechanistic interpretability have observed that neural network representations encode meaningful concepts as linear directions, often in superposition \citep{marks2024geometrytruthemergentlinear, nanda2023emergentlinearrepresentationsworld, jiang2024originslinearrepresentationslarge, park2025geometrycategoricalhierarchicalconcepts}.  Following \citet{elhage2022toymodelssuperposition} and \citet{park2024linearrepresentationhypothesisgeometry}, we primarily consider model representations $\mathbf{x}_p \in \mathbb{R}^{n_p}$ satisfying the following assumptions.

\begin{assumption}[Representation Assumptions]
\label{ass:lrh}
There exists a \emph{feature function} $\mathbf{x}: \mathcal{X} \to \mathbb{R}^n$ 
and a \emph{feature matrix} $W_p \in \mathbb{R}^{n_p \times n}$ such that:
\begin{enumerate}
    \item \textbf{Linear Decomposition}: For all $s \sim \mathcal{D}$,
    \[\mathbf{x}_p(s) = W_p \mathbf{x}(s).\]

    \item \textbf{Unit Norm}: The feature matrix $W_p \in \mathbb{R}^{n_p \times n}$ has unit-norm columns:
    \[\|W_{p}[:,i]\|_2 = 1 \quad \forall i \in [n]\]
    
    \item \textbf{Non-negativity}: $\mathbf{x}(s) \in \mathbb{R}^n_+$.
    
    \item \textbf{Sparsity}: There exists $S \in [0,1]$ such that $\forall i \in [n]$, 
    \[\Pr_{s \sim \mathcal{D}}(x_i(s) = 0) \geq S.\]

\end{enumerate}
We refer to $\mathbf{x}(s)$ as the \emph{ground-truth features}, where each $x_i(s)$ 
represents the activation level of concept $i$ for input $s$. Researchers typically assume each component $x_i$ of $\mathbf{x}$ corresponds to a human-interpretable concept \citep{bricken2023monosemanticity, cunningham2023sparseautoencodershighlyinterpretable, elhage2022toymodelssuperposition}.
\end{assumption}


\subsection{General Optimization Framework for SDL}


Now we formalize SDL as an optimization problem under the Representation Assumptions (Assumption~\ref{ass:lrh}). We follow previous approaches \citet{cui2025theoreticalunderstandingidentifiablesparse,elhage2022toymodelssuperposition} to omit the bias terms for mathematical simplicity.

\begin{definition}[Sparse Dictionary Learning]
\label{def:sdl}
A SDL model maps an input representation $\mathbf{x}_p(s) \in \mathbb{R}^{n_p}$ to a target representation $\mathbf{x}_r(s) \in \mathbb{R}^{n_r}$ through a two-layer architecture:
\begin{enumerate}[label=(\roman*)]
    \item An \emph{encoder} layer that maps $\mathbf{x}_p(s)$ to a latent space:
    \begin{equation}
    \label{eq:encoder}
    \mathbf{x}_q(s) = \sigma(W_E \mathbf{x}_p(s))
    \end{equation}
    
    \item A \emph{decoder} layer that maps the latents to $\mathbf{x}_r(s)$:
    \begin{equation}
    \label{eq:decoder}
    \hat{\mathbf{x}}_r(s) = W_D \mathbf{x}_q(s)
    \end{equation}
\end{enumerate}
where $\mathbf{x}_q(s) \in \mathbb{R}^{n_q}$, $W_E \in \mathbb{R}^{n_q \times n_p}$, $W_D \in \mathbb{R}^{n_r \times n_q}$, and $\sigma: \mathbb{R}^{n_q} \to \mathbb{R}^{n_q}$ is a sparsity-inducing activation function.

The SDL objective minimizes mean square error:
\begin{equation}
\label{eq:loss_basic}
\mathcal{L}_{\text{SDL}} = \mathbb{E}_{s \sim \mathcal{D}}\left[\|\mathbf{x}_r(s) - W_D \sigma(W_E \mathbf{x}_p(s))\|_2^2\right]
\end{equation}
\end{definition}

Under the Linear Representation Hypothesis (Assumption~\ref{ass:lrh}), both representations admit linear decompositions $\mathbf{x}_p(s) = W_p \mathbf{x}(s)$ and $\mathbf{x}_r(s) = W_r \mathbf{x}(s)$ in terms of ground-truth features $\mathbf{x}(s)$. The loss can thus be expressed as:
\begin{equation}
\label{eq:loss_features}
\mathcal{L}_{\text{SDL}} = \mathbb{E}_{s \sim \mathcal{D}}\left[\|W_r \mathbf{x}(s) - W_D \sigma(W_E W_p \mathbf{x}(s))\|_2^2\right]
\end{equation}

\subsection{Instantiations: Existing SDL Methods}

We now demonstrate how existing SDL methods instantiate our framework through adopting different choices of input-target pairs $(\mathbf{x}_p, \mathbf{x}_r)$ and activation functions $\sigma$, and proposing variants on the loss function $\mathcal{L}_{\text{SDL}}$ (See Appendix~\ref{app:taxonomy}).

\begin{figure}[H]
    \centering
    \includegraphics[width=0.7\linewidth]{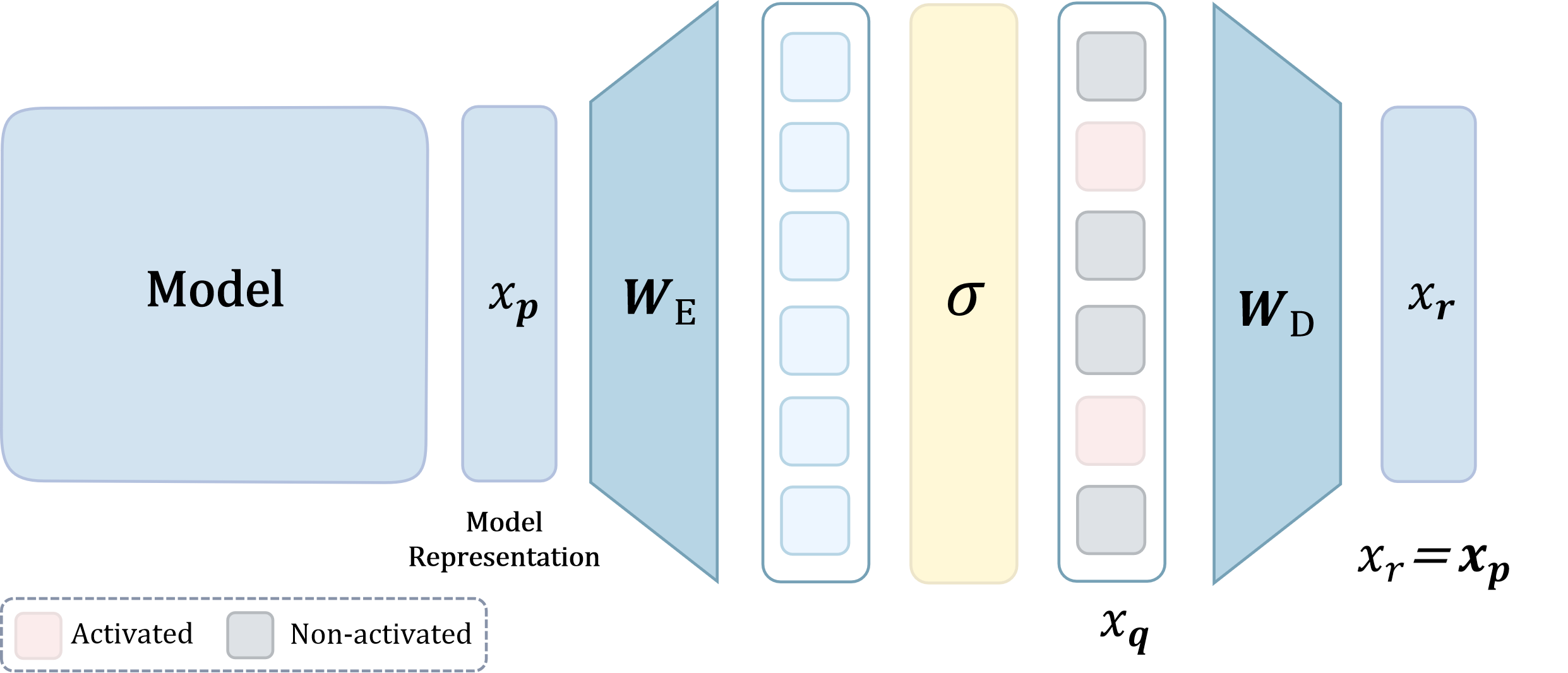}
    
    \caption{Sparse Autoencoder: encoder $W_E$ maps $\mathbf{x}_p$ to sparse latents $\mathbf{x}_q$, decoder $W_D$ reconstructs from $\mathbf{x}_q$.}
    \label{fig:illustration_sae}
\end{figure}

\textbf{Sparse Autoencoders (SAEs).} SAEs \citep{cunningham2023sparseautoencodershighlyinterpretable} decompose polysemantic activations into monosemantic components through sparsity constraints. In our framework, SAEs are characterized by setting $\mathbf{x}_r = \mathbf{x}_p$ (self-reconstruction). The encoder projects to a higher-dimensional sparse latent space, encouraging $\mathbf{x}_q(s)$ to capture the underlying ground-truth features $\mathbf{x}(s)$ (Figure~\ref{fig:illustration_sae}).

\begin{figure}[H]
    \centering
    \includegraphics[width=0.7\linewidth]{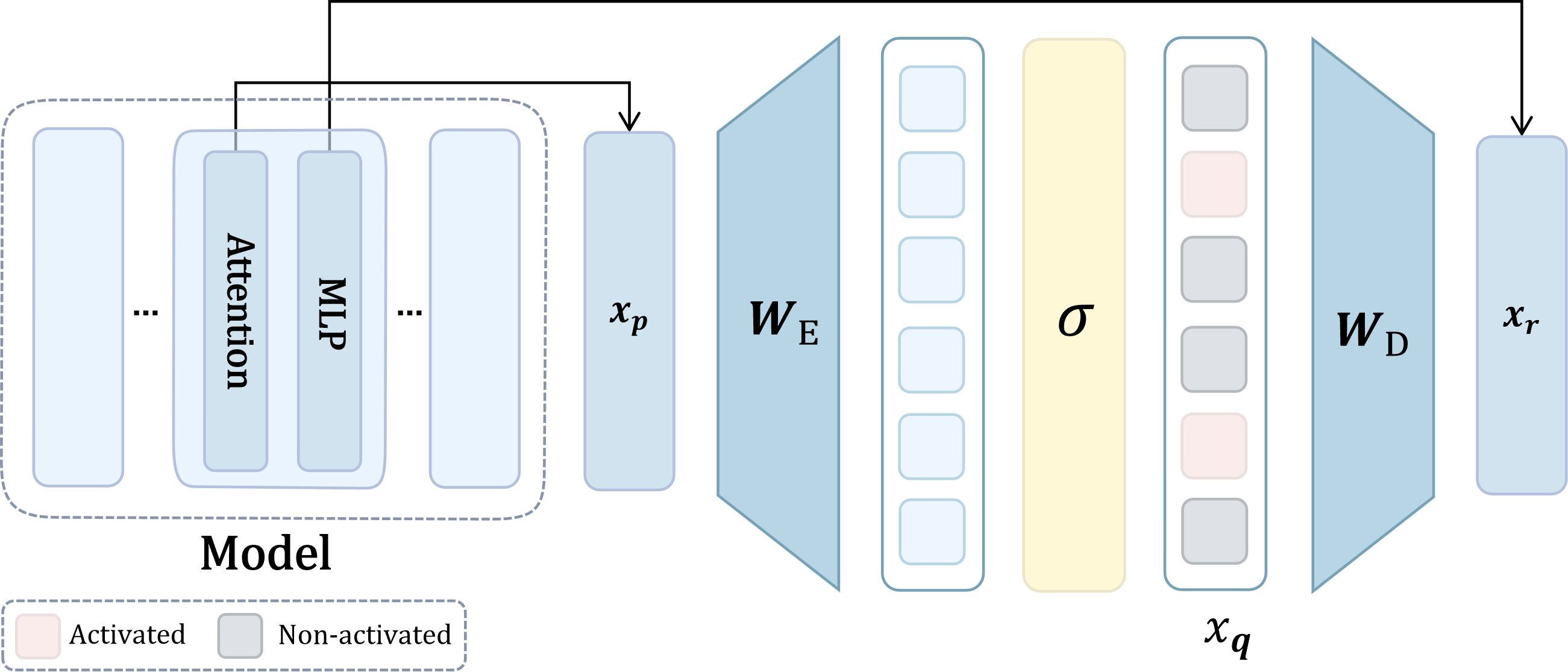}
    \caption{Transcoder: encoder $W_E$ maps $\mathbf{x}_{p}(s)$ to sparse latents $\mathbf{x}_q(s)$, decoder $W_D$ gives $\mathbf{x}_{r}(s)$ as a prediction of MLP's output.}
    \label{fig:illustration_transcoder}
\end{figure}

\textbf{Transcoders.} Transcoders \citep{dunefsky2024transcodersinterpretablellmfeature, paulo2025transcodersbeatsparseautoencoders} capture interpretable features in layer-to-layer transformations. Unlike SAEs, transcoders approximate the input-output function of a target component, such as a MLP, using a sparse bottleneck. In our proposed theoretical framework, transcoders set $\mathbf{x}_p = \mathbf{x}_{\text{mid}}(s)$ and $\mathbf{x}_r = \mathbf{x}_{\text{pre}}(s)$, where $\mathbf{x}_{\text{mid}}(s)$ denotes the inputs of one MLP block, and $\mathbf{x}_{\text{pre}}(s)$ denotes the prediction of MLP's outputs (Figure~\ref{fig:illustration_transcoder}).

\begin{figure}[H]
    \centering
    \includegraphics[width=0.7\linewidth]{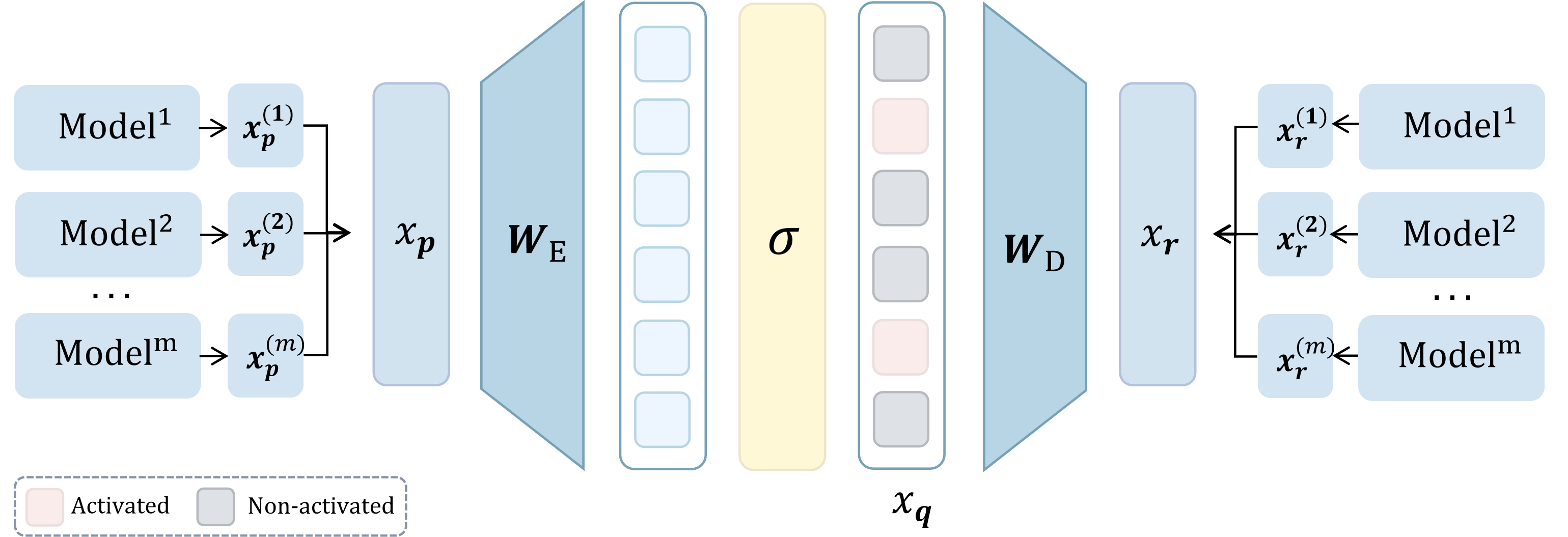}
    \caption{Crosscoder: encoder $W_E$ maps concatenated multi-layer input $\mathbf{x}_p$ to $\mathbf{x}_q$, decoder $W_D$ reconstructs multi-layer output $\mathbf{x}_r$.}
    \label{fig:illustration_crosscoder}
\end{figure}

\textbf{Crosscoders.} Crosscoders \citep{lindsey2024crosscoders} discover shared features across multiple representation sources by jointly encoding and reconstructing concatenated representations. In our framework, crosscoders set $\mathbf{x}_p = [\mathbf{x}_p^{(1)}; \ldots; \mathbf{x}_p^{(m)}]$ and $\mathbf{x}_r = [\mathbf{x}_r^{(1)}; \ldots; \mathbf{x}_r^{(m)}]$ where each superscript denotes a different source (Figure~\ref{fig:illustration_crosscoder}).

\textbf{Variants of SDL Methods.}
Various SDL methods fit into our theoretical framework but differ in their choices of activation functions and loss designs. \citet{bricken2023monosemanticity} and \citet{templeton2024scaling} use ReLU activation $\sigma_\text{ReLU}(z) = \max(0, z)$ with $L_1$ regularization on latents:
\begin{equation}
\mathcal{L} = \mathbb{E}_{s \sim \mathcal{D}}\left[\|\mathbf{x}_r(s) - W_D \sigma_\text{ReLU}(W_E \mathbf{x}_p(s))\|_2^2 + \lambda \|\mathbf{x}_q(s)\|_1\right]
\end{equation}
\citet{rajamanoharan2024jumpingaheadimprovingreconstruction} propose to utilize JumpReLU as the activation function:
\begin{equation}
\sigma_{\text{JumpReLU}}(z) = z \cdot H(z - \theta)
\end{equation}
combined with a smoothed Heaviside function to directly penalize the $L_0$ norm. \citet{makhzani2014ksparseautoencoders} introduce Top-$k$ activation:
\begin{equation}
\sigma_{\text{Top-}k}(z)_i = \begin{cases} z_i & \text{if } z_i \text{ is among the } k \text{ largest components} \\ 0 & \text{otherwise} \end{cases}
\end{equation}
\citet{bussmann2024batchtopksparseautoencoders} extend this to Batch Top-$k$ activation, which applies Top-$k$ selection across a batch to allow different samples to have different numbers of activated features. \citet{rajamanoharan2024improvingdictionarylearninggated} use ReLU activation with an additional Heaviside gating function. \citet{gao2024scalingevaluatingsparseautoencoders} employ Top-$k$ activation and introduce an auxiliary loss using Top-$k_{\text{aux}}$ dead latents:

\begin{equation}
\begin{aligned}
\mathcal{L} = \mathbb{E}_{s \sim \mathcal{D}}\Big[\|\mathbf{x}_r(s) - W_D \sigma_{\text{Top-}k}(W_E \mathbf{x}_p(s))\|_2^2 \\
+ \lambda_{\text{aux}} \|\mathbf{x}_p(s) - W_D' \sigma_{\text{Top-}k_{\text{aux}}}(W_E \mathbf{x}_p(s))\|_2^2\Big]
\end{aligned}
\end{equation}

to resurrect dead neurons. \citet{bussmann2025learningmultilevelfeaturesmatryoshka} and \citet{tang2025doesmodelfailautomatic} use multiple $k$ values with a multi-scale loss:
\begin{equation}
\mathcal{L} = \sum_{i=1}^{m} \lambda_i \mathbb{E}_{s \sim \mathcal{D}}\left[\|\mathbf{x}_r(s) - W_D \sigma_{\text{Top-}k_i}(W_E \mathbf{x}_p(s))\|_2^2\right]
\end{equation}
that sums reconstruction errors of different sparsity levels.

The variants described above demonstrate that diverse SDL methods can be unified under our general framework (Definition~\ref{def:sdl}) through specific choices of activation functions $\sigma$ and loss modifications. Critically, this unification enables subsequent theoretical analysis to apply to the entire family of SDL methods, rather than just a single architecture. We also state a key property for SDL activation functions:

\begin{equation}\label{eq:activation-property}
    \sigma(z)_i \in \{0, z_i\} \quad \forall i \in [n_q]
\end{equation}
This property holds for ReLU, JumpReLU, Top-$k$, Batch Top-$k$, and their compositions, the primary sparsity mechanisms used in practice. 
\section{Theoretical Results}
\label{sec:theoretical_results}



Despite the empirical success of SDL methods across diverse applications, a significant gap exists between their practical use and our theoretical understanding of their optimization dynamics. This gap has important consequences: practitioners employ techniques like dead neuron resampling \citep{bricken2023monosemanticity} and observe phenomena like feature absorption \citep{chanin2025absorptionstudyingfeaturesplitting} without rigorous explanations for why these occur or how to systematically address them. Without theoretical grounding, the development of improved SDL methods remains largely empirical, potentially leaving highly effective techniques underexplored.

Our theoretical analysis bridges this gap by characterizing the SDL optimization landscape under the Linear Representation Hypothesis. Section~\ref{subsec:approximate} establishes a loss approximation enabling tractable analysis (Theorem~\ref{thm:loss-decomposition}). Section~\ref{subsec:biconvex} proves SDL exhibits piecewise biconvex structure within activation pattern regions (Theorem~\ref{thm:biconvex}). Section~\ref{subsec:global-minimum} characterizes the global minimum and shows the optimization is underdetermined  (Theorem~\ref{thm:global-existence} and ~\ref{thm:zero-loss-conditions}). Section~\ref{subsec:local-minimum} establishes the prevalence of spurious partial minima exhibiting polysemanticity (Theorem~\ref{thm:local-existence}). Section~\ref{subsec:feature-absorption} shows that hierarchical concept structures naturally induce feature absorption patterns that manifest as partial minima (Theorem~\ref{thm:hierarchical-absorption}). We provide full proofs in Appendix~\ref{app:proofs}.

\subsection{Approximate Loss with Feature Reconstruction}
\label{subsec:approximate}

We establish the bound for $S$ satisfying $(1-S) \leq \frac{1}{n}$, the ${\mathcal{L}}_{\text{SDL}}$ can be approximated by per-feature reconstruction.

\begin{theorem}[Loss Approximation]\label{thm:loss-decomposition}
Under Assumption~\ref{ass:lrh}, let 
$C = \sup_{s \sim \mathcal{D}} \|\mathbf{x}_r(s) - W_D\sigma(W_E\mathbf{x}_p(s))\|_2^2 < \infty$.
Define the approximate loss as:
\begin{equation}
\tilde{\mathcal{L}}_{\text{SDL}}(W_D, W_E) 
    := \sum_{d=1}^n M_d \left\| w_r^d - W_D \sigma(W_E w_p^d) \right\|^2
\end{equation}
where $M_d = \Pr(\mathbf{x}(s) = x_d(s)e_d) \cdot \mathbb{E}[x_d(s)^2 \mid x_d(s) > 0]$.
Then for any $S$ satisfying $(1 - S) \leq \frac{1}{n}$:
\begin{equation}
\left|\mathcal{L}_{\text{SDL}} - \tilde{\mathcal{L}}_{\text{SDL}}\right| 
    \leq n^2 C (1-S)^2
\end{equation}
\end{theorem}

This approximation isolates the contribution of each ground-truth feature, making the optimization landscape amenable to analysis techniques applied subsequently.

\subsection{SDL Optimization is Piecewise Biconvex}
\label{subsec:biconvex}

While the SDL loss is non-convex globally due to activation discontinuities, it exhibits favorable convex structure within regions of fixed activation patterns.

\begin{theorem}[Bi-convex Structure of SDL]\label{thm:biconvex}
Consider the approximate loss $\tilde{\mathcal{L}}_{\text{SDL}}(W_D, W_E)$.
For an activation pattern $\mathcal{P} = (\mathcal{F}_1, \ldots, \mathcal{F}_{n_q})$ 
(See Definition~\ref{def:activation-pattern}), define the corresponding 
activation pattern region as:
\begin{align}
    \Omega_{\mathcal{P}} = \Bigl\{ &W_E \in \mathbb{R}^{n_q \times n_p} \;\Big|\; 
    \forall i \in [n_q], \nonumber\\
    &\mathcal{F}_i = \bigl\{ d \in [n] : 
    \bigl(\sigma(W_E w_p^d)\bigr)_i > 0 \bigr\} 
    \Bigr\},
\end{align}
Then $\tilde{\mathcal{L}}_{\text{SDL}}$ exhibits bi-convex structure over 
$\mathbb{R}^{n_r \times n_q} \times \Omega_{\mathcal{P}}$:
\begin{enumerate}
    \item For any fixed $W_E \in \Omega_{\mathcal{P}}$, $W_D \mapsto 
    \tilde{\mathcal{L}}_{\text{SDL}}(W_D, W_E)$ is convex in $W_D$.
    \item For any fixed $W_D \in \mathbb{R}^{n_r \times n_q}$, $W_E \mapsto 
    \tilde{\mathcal{L}}_{\text{SDL}}(W_D, W_E)$ is convex in $W_E$ over $\Omega_{\mathcal{P}}$.
\end{enumerate}
\end{theorem}

This establishes SDL as a biconvex optimization problem within each activation pattern, bridging mechanistic interpretability with classical biconvex optimization theory, enabling both theoretical analysis of the optimization landscape and implementation of biconvex methods for SDL.

\subsection{Characterizing the Global Minimum of SDL}
\label{subsec:global-minimum}

Characterizing the global minimum reveals both the success conditions for SDL training and the fundamental underdetermined nature of the optimization problem.

\begin{theorem}[Successful Reconstruction Achieves Near-Zero Loss]\label{thm:global-existence}
Consider the approximate SDL loss
\begin{equation}
    \tilde{\mathcal{L}}_{\text{SDL}}(W_D, W_E) = \sum_{d=1}^n M_d 
    \left\| w_r^d - W_D \sigma(W_E w_p^d) \right\|^2
\end{equation}
where $M_d = \Pr(\mathbf{x}(s) = x_d(s)e_d) \cdot \mathbb{E}[x_d(s)^2 \mid x_d(s) > 0]$,
and let $M = \max_{i \neq j} \langle w_p^i, w_p^j \rangle$ denote the maximum interference.
When $n_q \geq n$ and $\sigma$ satisfies $\sigma(z)_i \in \{0, z_i\}$ for all $i$, 
the configuration
\begin{equation}
\label{equation:global_minima}
    W_D^* = [W_r, \mathbf{0}], \quad W_E^* = \begin{bmatrix} W_p^\top \\ \mathbf{0} \end{bmatrix}
\end{equation}
where $\mathbf{0}$ denotes zero padding to dimension $n_q$, satisfies:
\begin{equation}
    \tilde{\mathcal{L}}_{\text{SDL}}(W_D^*, W_E^*) 
    \leq n^2 M^2 \sum_{d=1}^{n} M_d
\end{equation}
\end{theorem}

\begin{theorem}[Necessary and Sufficient Conditions for Zero Loss]\label{thm:zero-loss-conditions}
The approximate loss satisfies $\tilde{\mathcal{L}}_{\text{SDL}}(W_D, W_E) = 0$ if and only if
\begin{equation}
    w_r^d = W_D \sigma(W_E w_p^d) \quad \text{for all } d \in [n]
\end{equation}
\end{theorem}

\begin{figure*}
    \centering
    \includegraphics[width=0.9\linewidth]{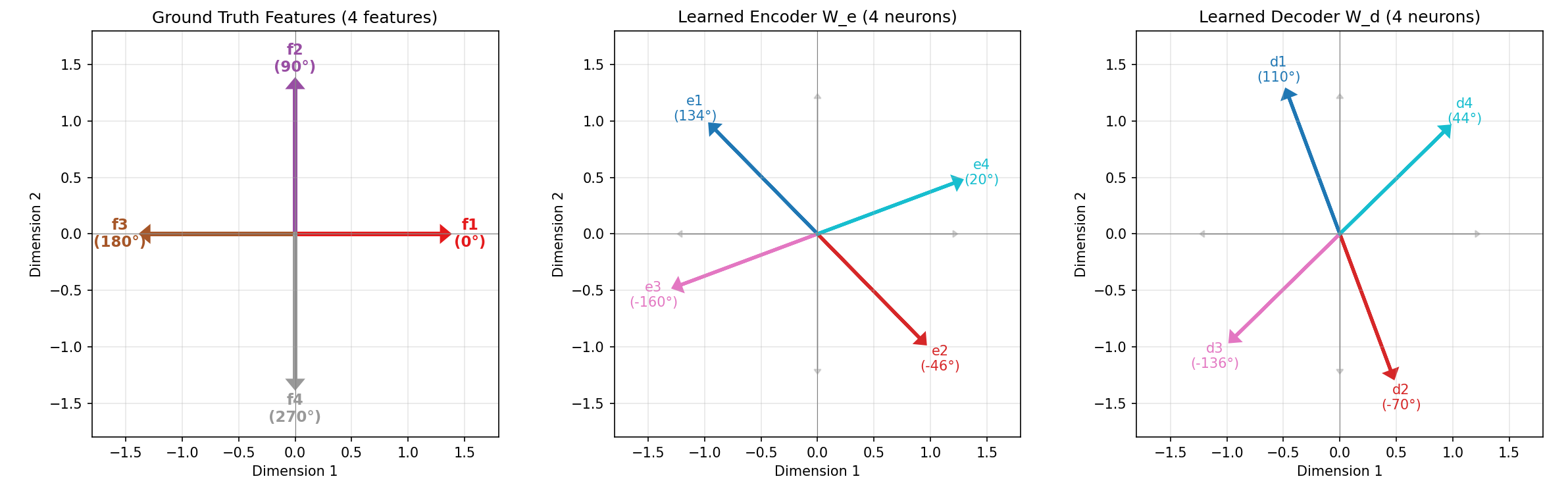}
    \caption{\textbf{Zero reconstruction loss without recovering ground-truth features.} We design the Linear Representation Bench that enable full knowledge of the ground truth features to study SDL methods. We observe one concerning phenomenon that these methods can achieve zero loss without recovering any ground truth features.  \textbf{Left:} Four ground-truth feature directions. \textbf{Middle:} Learned encoder directions fail to align with ground truth. \textbf{Right:} Learned decoder directions are rotated accordingly. Although $\tilde{\mathcal{L}}_{\text{SDL}} \approx 0$, the learned features bear no correspondence to interpretable ground-truth concepts, demonstrating the underdetermined nature of SDL optimization.}
    \label{fig:zero-recovery}
\end{figure*}

Theorem~\ref{thm:global-existence} provides a constructive global minimum that recovers ground-truth features, while Theorem~\ref{thm:zero-loss-conditions} reveals the complete solution space. This system of $n$ vector equations is underdetermined when $n_q > n$, admitting multiple solutions beyond the feature-recovering configuration. Critically, some solutions achieve zero reconstruction loss without recovering any ground-truth features (Figure~\ref{fig:zero-recovery}). 


\subsection{Characterizing Spurious Partial Minima of SDL}
\label{subsec:local-minimum}

Beyond the global minimum, SDL optimization exhibits spurious partial minima where neurons exhibit polysemanticity—responding to multiple unrelated features.

\begin{example}[Spurious Partial Minimum]\label{ex:partial-minimum}
Consider $n = n_p = n_q = n_r = 2$, $\sigma=\sigma_{\text{Top-1}} \cdot \sigma_{\text{ReLU}}$, $S \to 1$, and $M_1 = M_2 = 1$. Let:
\begin{equation}
w_p^1 = \begin{bmatrix}1 \\ 0\end{bmatrix}, \quad w_p^2 = \begin{bmatrix}0 \\ 1\end{bmatrix}, \quad 
w_r^1 = \begin{bmatrix}1 \\ 0\end{bmatrix}, \quad w_r^2 = \begin{bmatrix}0 \\ 1\end{bmatrix}
\end{equation}

Consider the configuration where neuron 1 activates for both features while neuron 2 remains dead:
\begin{equation}\label{eq:partial-explicit}
W_E^* = \begin{bmatrix}
1 & 1 \\
0 & 0
\end{bmatrix}, \quad W_D^* = \begin{bmatrix}
1/2 & 0 \\
1/2 & 0
\end{bmatrix}
\end{equation}

Within the activation pattern region $\Omega_{\mathcal{A}} = \{W_E : \langle w_E^1, w_p^1\rangle > 0, \langle w_E^1, w_p^2\rangle > 0\}$, direct calculation shows $\nabla_{W_D} \tilde{\mathcal{L}} = 0$ and $\nabla_{W_E} \tilde{\mathcal{L}} = 0$. By traditional biconvex optimization theory \citep{gorski2007biconvex}, $(W_D^*, W_E^*)$ is a partial optimum. However, this configuration exhibits polysemanticity with suboptimal loss: $\tilde{\mathcal{L}}_{\text{SDL}}(W_D^*, W_E^*) = 1 > 0$.
\end{example}

\begin{definition}[Activation Pattern]\label{def:activation-pattern}
An \emph{activation pattern} is a collection $\mathcal{P} = (\mathcal{F}_1, \ldots, \mathcal{F}_{N})$ where $\mathcal{F}_i \subseteq [n]$ denotes the set of ground-truth features that activate neuron $i$.

An activation pattern $\mathcal{P}$ is called:
\begin{itemize}
    \item \emph{Polysemantic} if $\exists i \in [N]$ such that $|\mathcal{F}_i| \geq 2$ (at least one neuron responds to multiple features).
    \item \emph{Realizable} if there exists an encoder $\sigma(W_E(\cdot)) \in \mathbb{R}^{N \times n_p}$ such that:
    \begin{equation}
        \forall i \in [N], \quad \mathcal{F}_i = \left\{d \in [n] : \left(\sigma(W_E w_p^d)\right)_i > 0\right\}
    \end{equation}
    That is, neuron $i$ activates for exactly the features in $\mathcal{F}_i$ when they appear in isolation.
\end{itemize}
\end{definition}

\begin{theorem}[Prevalence of Spurious Partial Minima]\label{thm:local-existence}
Under Assumptions~\ref{ass:lrh} with $n \geq 2$ and $n_q \geq n$, for any activation pattern $\mathcal{P} = (\mathcal{F}_1, \ldots, \mathcal{F}_{n_q})$ that is realizable, polysemantic, and forms a partition of $[n]$, there exists a partial minimum $(W_D^*, W_E^*)$ of $\tilde{\mathcal{L}}_{\text{SDL}}$ exhibiting this pattern with positive loss.
\end{theorem}

This establishes that partial minima are pervasive in SDL optimization: every realizable polysemantic activation pattern corresponds to a partial optima point where gradient descent can become trapped, a persistent challenge for SDL.


\subsection{Theoretical Explanation for Feature Absorption}
\label{subsec:feature-absorption}

Feature absorption—where one neuron captures, or "absorbs", a specific sub-concept while another responds to remaining related concepts—frequently occurs in SDL training~\citep{chanin2025absorptionstudyingfeaturesplitting}. Though prevalently encountered and unwanted, researchers have limited understanding about why feature absorption occurs in SDL training. Here we show hierarchical concept structures naturally introduce realizable activation patterns exhibiting feature absorption and therefore connected with the framework's partial minima.

\begin{figure*}
    \centering
    \includegraphics[width=0.9\linewidth]{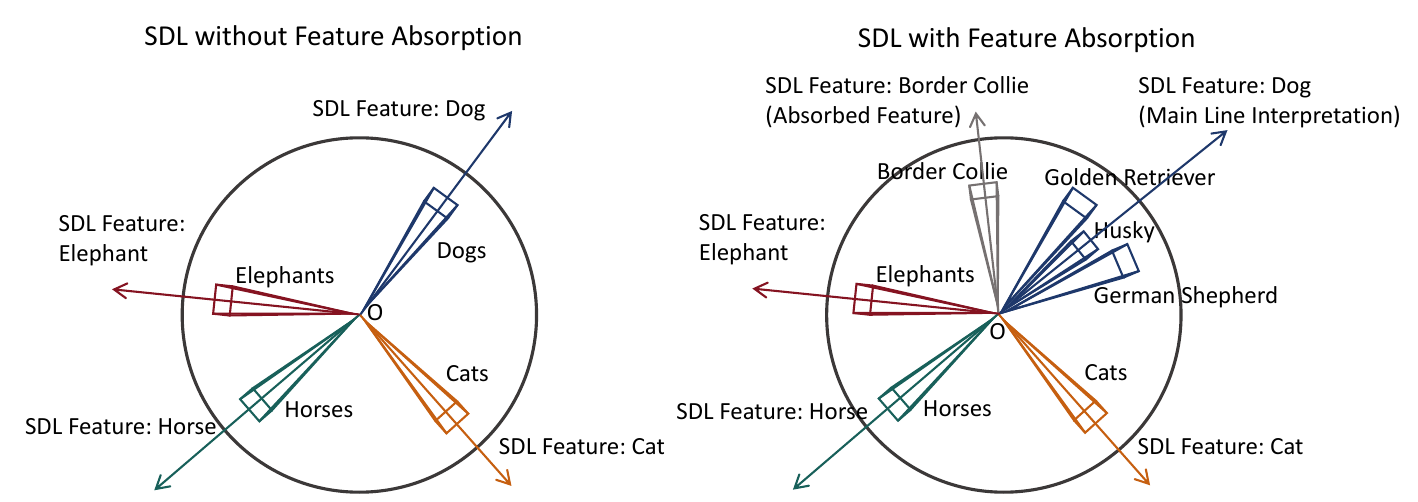}
    \caption{\textbf{Feature absorption emerges from hierarchical concept structure.} Left: Ideal SDL features without absorption. Right: hierarchical concept structure exists and only a proportion of the sub-concepts of "Dog" can activate the SDL feature.}
    \label{fig:feature_absorption}
\end{figure*}

\begin{example}[Feature Absorption]
\label{ex:dog-absorption}
Consider a representation space with a parent concept "Dog" and four sub-concepts: "Border Collie", "Golden Retriever", "Husky", and "German Shepherd". Ideally, SDL learns separate monosemantic neurons for Dog, Cat, Horse, and Elephant, with each dog breed activating only the Dog neuron. However, feature absorption can result in the pattern in Figure~\ref{fig:feature_absorption}: one neuron exclusively captures "Border Collie" (absorbed feature), while another responds to the remaining three breeds collectively (main line interpretation).
\end{example}

\begin{definition}[Hierarchical Concept Structure]
\label{def:hierarchical-features}
A set of ground-truth features exhibits \emph{hierarchical structure} if it is composed of a parent concept $p$ and a set of sub-concepts $c_1, \ldots, c_k$ satisfying: $p(x)>0 \iff \exists i\in[k], c_i(x)>0$.
\end{definition}

\begin{theorem}[Feature Absorption from Hierarchical Structure]
\label{thm:hierarchical-absorption}
Suppose there exist $M$ parent concepts with hierarchical decompositions into sub-concepts: for each $i \in [M]$, parent concept $d_i$ decomposes into sub-concepts $\mathcal{F}_i = \{d_{i,1}, \ldots, d_{i,k_i}\}$ where $k_i \geq 2$.

If the activation pattern $\mathcal{P} = (\mathcal{F}_1, \ldots, \mathcal{F}_M)$ is realizable, then $\forall i^* \in [M]$, $\exists j^* \in [k_{i^*}]$ such that the following activation pattern exhibiting feature absorption is realizable:
\begin{equation}
\mathcal{P}' = (\mathcal{F}_1, \ldots, \mathcal{F}_{i^*} \setminus \{d_{i^*,j^*}\},\{d_{i^*,j^*}\}, \ldots, \mathcal{F}_M)
\end{equation}
\end{theorem}

Theorem~\ref{thm:hierarchical-absorption} explains why feature absorption persists even with careful training: hierarchical concept structures naturally induce realizable polysemantic patterns that manifest as partial minima, providing theoretical grounding for this widely observed empirical phenomenon.

\section{Method}
\label{sec:method}
Our theoretical analysis reveals a fundamental challenge in SDL optimization: the underdetermined nature of the loss landscape (Theorem~\ref{thm:zero-loss-conditions}), and the solution space admits multiple configurations—some achieving zero reconstruction loss without recovering any interpretable ground-truth features (Figure~\ref{fig:zero-recovery}). These theoretical findings motivate our method, \textbf{feature anchoring}, a technique that constrains a subset of features to align with known semantic directions.

\subsection{Anchor Feature Extraction}

Feature anchoring requires identifying $k$ anchor features $\{\tilde{\mathbf{w}}_p^{(i)}, \tilde{\mathbf{w}}_r^{(i)}\}_{i=1}^k$ that represent semantically meaningful directions in the representation space. We present two methods for obtaining these anchors:

\paragraph{Ground-Truth Features (Linear Representation Bench).}
When ground-truth features are available—as in our Linear Representation Bench where features are known by construction—we directly use a random subset of $k$ ground-truth feature directions:
\begin{equation}
\tilde{w}_p^{(i)} = W_p^{\text{true}}[:, i], \quad \tilde{w}_r^{(i)} = W_r^{\text{true}}[:, i], \quad i \in \mathcal{K}
\end{equation}
where $\mathcal{K} \subset [n]$ is a randomly selected subset of size $k$.

\paragraph{Subpopulation Mean Embeddings.}
For real-world datasets where ground-truth features are unknown, we identify semantic subpopulations \citet{luo2024llmdatasetanalystsubpopulation} and compute their mean representations. Specifically, given a labeled dataset $\mathcal{D} = \{(s_j, y_j)\}_{j=1}^N$, where $y_j \in \{1, \ldots, C\}$, we compute the mean of representations for one subpopulation cluster, which is considered as a proxy for ground-truth features:
\begin{equation}
\bar{\mathbf{x}}_p^{(c)} = \frac{1}{|\{j : y_j = c\}|} \sum_{j: y_j = c} \mathbf{x}_p(s_j)
\end{equation}
Then we normalize each mean representation to obtain anchor directions:
\begin{equation}
\tilde{\mathbf{w}}_p^{(c)} = \frac{\bar{\mathbf{x}}_p^{(c)}}{\|\bar{\mathbf{x}}_p^{(c)}\|_2}
\end{equation}

\subsection{SDL Loss Function with Feature Anchoring}

Given $k$ anchor features $\{\tilde{\mathbf{w}}_p^{(i)}, \tilde{\mathbf{w}}_r^{(i)}\}_{i=1}^k$, we modify the SDL optimization objective to include an anchoring penalty that constrains the first $k$ encoder rows and decoder columns to align with these anchors.

The complete anchored SDL objective is:
\begin{equation}
\mathcal{L}_{\text{SDL-FA}} = \mathcal{L}_{\text{SDL}} + \lambda_{\text{anchor}} \mathcal{L}_{\text{anchor}}
\end{equation}
where $\mathcal{L}_{\text{SDL}}$ is the standard SDL loss (Equation~\ref{eq:loss_basic}) and the anchoring loss is:
\begin{equation}
\label{eq:anchor_loss}
\begin{split}
    \mathcal{L}_{\text{anchor}} &= \|W_E[1{:}k,:] - [\tilde{w}_p^{(1)}, \ldots, \tilde{w}_p^{(k)}]^\top\|^2_F  \\
    &\quad + \|W_D[:,1{:}k] - [\tilde{w}_r^{(1)}, \ldots, \tilde{w}_r^{(k)}]\|^2_F
\end{split}
\end{equation}

This feature anchoring technique (Equation~\ref{eq:anchor_loss}) reduces the underdetermined nature of SDL training, and is \emph{method-agnostic}: it applies equally to SAEs, transcoders, crosscoders, and their variants (TopK SAEs, Matryoshka SAEs, etc.) since it only constrains the encoder $W_E$ and decoder $W_D$ matrices that are common to all SDL architectures. We believe this universality makes feature anchoring a broadly useful technique for improving performance across SDL methods.
\section{Experimental Results}
\label{sec:experiments}

We first evaluate feature anchoring with various SDL methods on the Linear Representation Bench (Section~\ref{subsec:linear_rep_bench}), a synthetic benchmark with fully accessible ground-truth features that precisely instantiates our theoretical assumptions. Second, we validate feature anchoring on real-world data by training SDL methods on CLIP embeddings of ImageNet-1K. Third, we provide empirical evidence showing that dead neuron resampling helps escape spurious local minima in large language models. Ablation studies are in Appendix~\ref{app:ablation}.

\subsection{Results on The Linear Representation Bench}
\label{subsec:linear_rep_bench}

To validate our theoretical predictions, we design the Linear Representation Bench, a benchmark that precisely instantiates Assumptions~\ref{ass:lrh} with fully known GT features.

\textbf{Data Generation.} We generate synthetic representations $\mathbf{x}_p(s) = W_p^{\text{true}} \mathbf{x}(s)$ where ground-truth features $\mathbf{x}(s) \in \mathbb{R}^n$ follow shifted exponential distributions with sparsity $\mathcal{S}$. The feature matrix $W_p^{\text{true}} \in \mathbb{R}^{n_p \times n}$ is constructed via gradient-based optimization (Detailed in Appendix~\ref{app:bench}).

\textbf{Metrics.} Given learned features $W_p^{\text{learned}}=W_E^\top \in \mathbb{R}^{n_p \times n_q}$ and ground truth features $W_p^{\text{true}} \in \mathbb{R}^{n_p \times n}$ (columns are unit-norm feature directions), we compute the similarity matrix $\mathbf{S} = |W_p^{\text{learned}\top} W_p^{\text{true}}| \in \mathbb{R}^{n_q \times n}$. For each ground truth feature $i \in [n]$, we find its best match score: $s_i = \max_j S_{ji}$. 

\begin{itemize}
    \item \textbf{GT Recovery} is the fraction of ground truth features with $s_i > \tau$, $\mathbf{M_{GT}}(W_p^{\text{learned}})=\frac{1}{n}\sum_{i=1}^{n} \mathbf{1}\{s_i > \tau\}$.
    \item \textbf{Maximum Inner Product} is the mean of best match scores: $\mathbf{M_{IP}}(W_p^{learned}) = \frac{1}{n}\sum_{i=1}^{n} s_i$.
\end{itemize}

\begin{figure}[h]
    \centering
    \includegraphics[width=0.82\linewidth]{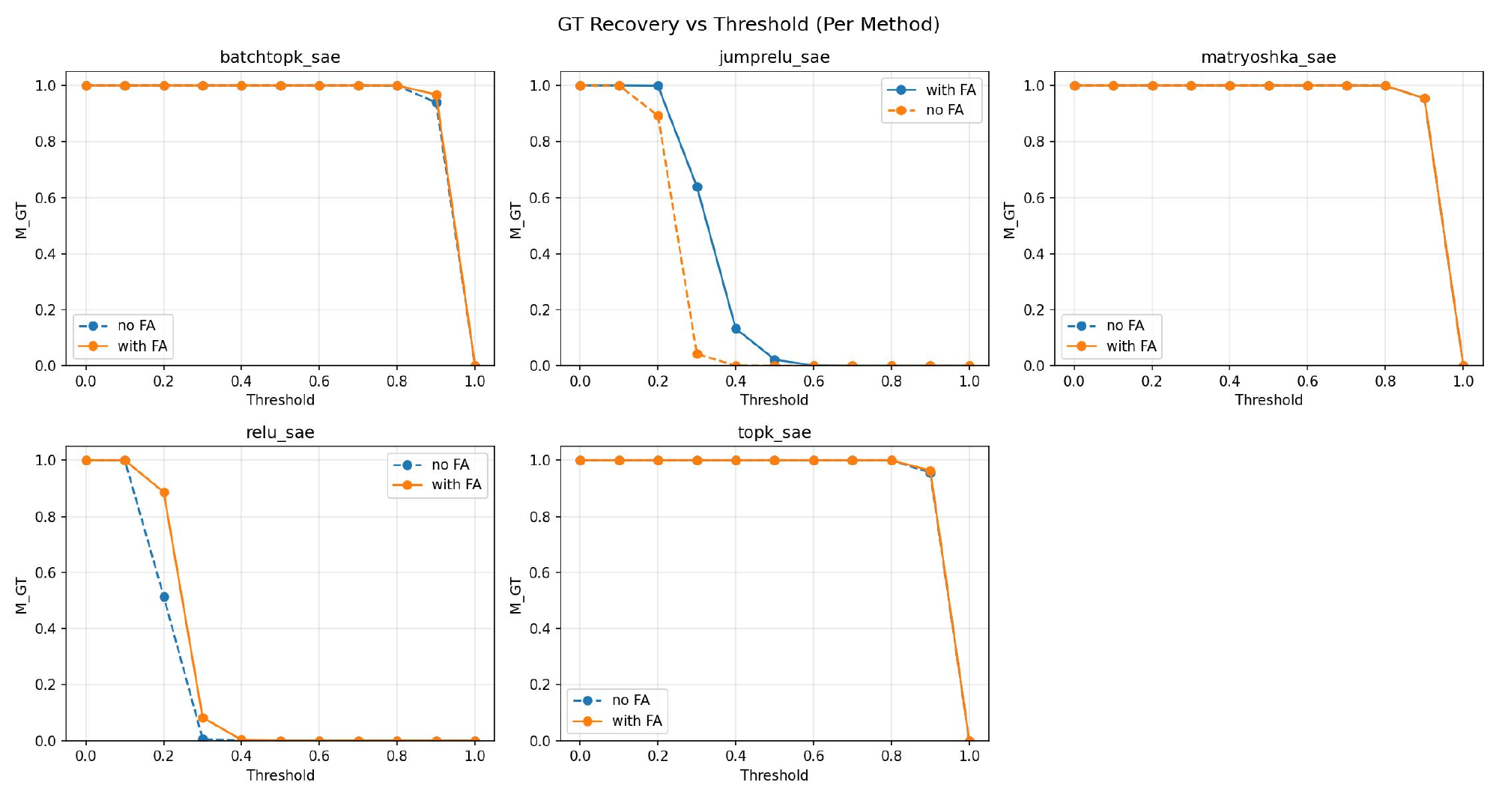}
    \vspace{-5mm}
    \caption{\textbf{$\mathbf{M_{GT}}$ vs. Threshold.} Across all SDL methods, feature anchoring consistently improves feature recovery, as measured by $\mathbf{M_{GT}}$ over a range of thresholds.}
    \label{fig:gt_curve}
\end{figure}

As shown in Table~\ref{tab:LRB_results} and Figure~\ref{fig:gt_curve}, feature anchoring significantly improves feature recovery performances.

\begin{table}
\centering
\caption{Feature recovery results on the Linear Representation Bench ($n=1000$, $n_p=n_r=768$, $n_q=16000$, $\mathcal{S}=0.99$). Evaluation is performed only on features not used for anchoring. Feature anchoring (FA) consistently improves both GT Recovery ($\mathbf{M_{GT}}$) and Maximum Inner Product ($\mathbf{M_{IP}}$) across all SDL methods. *: For some methods, $\mathbf{M_{GT}}$ is 0\% because all best-match similarities fall below the evaluation threshold (0.95). In these cases, $\mathbf{M_{IP}}$, which remains well-aligned with $\mathbf{M_{GT}}$, serves as a more informative indicator of recovery performance. See Figure~\ref{fig:gt_curve} for a threshold-dependent view of $\mathbf{M_{GT}}$.}

\label{tab:LRB_results}
\begin{tabular}{lcc}
\toprule
\textbf{Method} & $\mathbf{M_{GT}}$ $\uparrow$ & $\mathbf{M_{IP}}$ $\uparrow$\\
\midrule
ReLU SAE & 0.00\%* & 0.205 \\
\quad + Feature Anchoring & 0.00\%* & \textbf{0.246}\\
\midrule
JumpReLU SAE &  0.00\%* & 0.237\\
\quad + Feature Anchoring &  0.00\%* & \textbf{0.333}\\
\midrule
TopK SAE & 84.90\% & 0.983 \\
\quad + Feature Anchoring & \textbf{87.63\%} & \textbf{0.986} \\
\midrule
BatchTopK SAE & 84.80\% & 0.981\\
\quad + Feature Anchoring & \textbf{89.38\%} &\textbf{ 0.988 }\\
\midrule
Matryoshka SAE & 83.70\%& 0.982\\
\quad + Feature Anchoring & \textbf{87.32\%} & \textbf{ 0.985} \\
\midrule
Transcoder & 23.60\% & 0.838  \\
\quad + Feature Anchoring & \textbf{25.05\%} & 0.838\\
\midrule
Crosscoder & 56.42\% &  0.940\\
\quad + Feature Anchoring & \textbf{57.71\%} & \textbf{0.941}\\
\bottomrule
\end{tabular}
\end{table}

\subsection{Results on CLIP Embeddings of ImageNet}

To validate that feature anchoring generalizes beyond synthetic benchmarks to real-world representations, we apply our method to CLIP embeddings of ImageNet-1K \citep{russakovsky2015imagenetlargescalevisual}. We extract CLIP-ViT-B/32 \citep{radford2021learningtransferablevisualmodels} image embeddings for all ImageNet training images and compute ground-truth anchor features as normalized class mean embeddings: $\tilde{w}_p^{(c)} = \bar{x}_p^{(c)} / \|\bar{x}_p^{(c)}\|_2$ where $\bar{x}_p^{(c)}$ is the mean embedding over all images in class $c$. We train TopK, BatchTopK, and Matryoshka SAEs with $n_p = n_r = 768$, $n_q = 16,384$, and use $k = 30$ anchors with $\lambda_{\text{anchor}} = 1.0$. Table \ref{tab:clip_results} shows feature anchoring consistently improves feature recovery (Examples in Appendix~\ref{sec:qualitative_examples}).

\begin{table}[H]
\centering
\caption{Feature recovery on CLIP embeddings. As ground-truth features are not explicitly available in this setting, we use subpopulation mean embeddings as a proxy for the underlying feature directions. *: 0\% recovery resulted from the same issue as Talble~\ref{tab:LRB_results}.}
\label{tab:clip_results}
\begin{tabular}{lcc}
\toprule
\textbf{Method} & $\mathbf{M_{GT}}$ $\uparrow$ & $\mathbf{M_{IP}}$ $\uparrow$\\
\midrule
TopK SAE & 0.00\%* & 0.517\\
\quad + Feature Anchoring & \textbf{24.13\%} & \textbf{0.851}\\
\midrule
BatchTopK SAE & 0.00\%* &  0.508\\
\quad + Feature Anchoring & \textbf{24.13\%} &\textbf{0.847}\\
\midrule
Matryoshka SAE & 0.00\%* & 0.683\\
\quad + Feature Anchoring & \textbf{24.45\%} & \textbf{0.858} \\
\bottomrule
\end{tabular}
\end{table}

\vspace{-5mm}
\subsection{Neuron Resampling Helps Escape Partial Minima}
Our partial minima analysis (Theorem~\ref{thm:local-existence}) proves the prevalence of spurious partial minima and connects them to dead neurons ($\mathcal{F}_i = \emptyset$). To address this, we utilize neuron resampling \citep{bricken2023monosemanticity} to help SDL training overcome these partial minima. We argue that resampling reinitializes dead neurons toward under-reconstructed directions, perturbing the optimization away from the partial minima. 

\begin{figure}[H]
    \centering
    \includegraphics[width=0.4\linewidth]{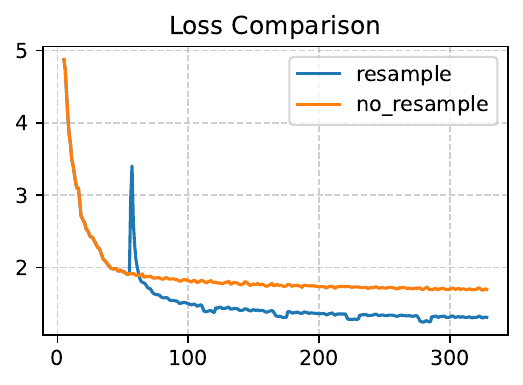}
    \vspace{-5mm}
    \caption{\textbf{Feature resampling accelerates convergence and improves final loss.} Training curves on Llama 3.1 8B comparing standard SAE training (blue) with periodic dead neuron resampling after 5000 steps (in the plot x axis is scaled by 100).}
    \label{fig:resampling}
\end{figure}

To validate this, we train SAEs on Llama 3.1 8B Instruct (layer 12, dimension 4096) with latent dimension 131,072 for 30,000 steps on FineWeb-Edu \citep{penedo2024finewebdatasetsdecantingweb}, processing 4096 token activations per step. We compare standard training against feature resampling after 5000 steps. As shown in Figure~\ref{fig:resampling}, resampling enables the optimizer to escape spurious local minima, achieving lower final loss.

\vspace{-5mm}
\vspace{-4mm}

\section{Conclusion}
\label{sec:conclusion}

We develop the first unified theoretical framework for Sparse Dictionary Learning in mechanistic interpretability, demonstrating how diverse SDL methods instantiate a single optimization problem. We prove that SDL exhibits piecewise biconvex structure, bridging mechanistic interpretability with classical optimization theory. We characterize the global minimum and establish that the optimization is fundamentally underdetermined, admitting solutions that achieve zero reconstruction loss without recovering interpretable features. We demonstrate that spurious partial minima exhibiting polysemanticity are pervasive, and prove that hierarchical concept structures naturally induce feature absorption patterns that manifest as partial minima. To validate our theory, we design the Linear Representation Bench with fully accessible ground-truth features, and propose feature anchoring—a technique applicable to all SDL methods that addresses the underdetermined nature of optimization. 


\section*{Impact Statement}
This paper presents work whose goal is to advance the field of Machine Learning. There are many potential societal consequences of our work, none which we feel must be specifically highlighted here.

\section*{Code Availability}
We will provide full scripts of the Linear Representation Bench and feature anchoring upon acceptance.

\section*{Declaration of LLM Usage}
We admit the usage of Claude Sonnet-4.6 for coding and polishing the manuscript. 

\bibliography{main}
\bibliographystyle{tmlr}

\appendix
\appendix
\onecolumn

\clearpage 

\clearpage 

\section{Taxonomy for Sparse Dictionary Learning in Mechanistic Interpretability}
\label{app:taxonomy}


\begin{figure}[!ht]
    \centering
    \includegraphics[height=0.87\textheight]{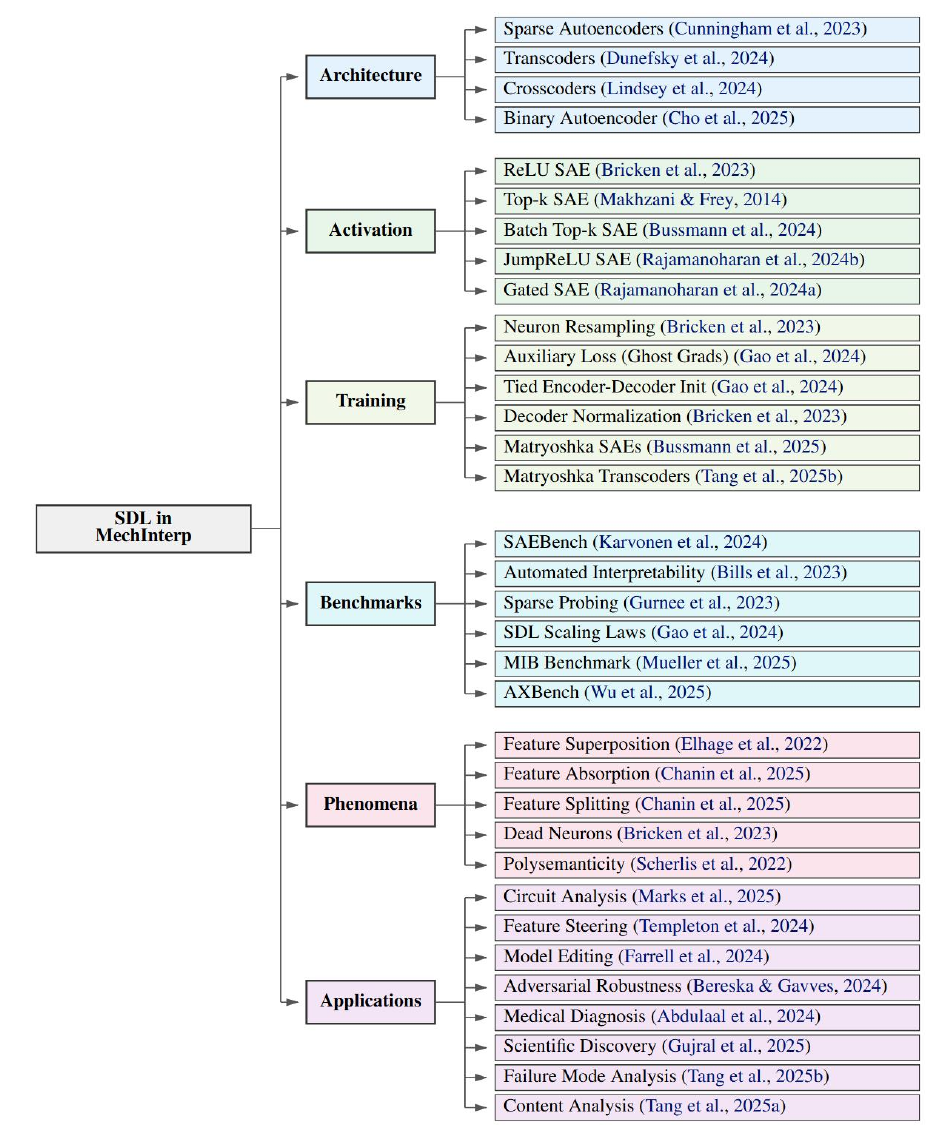}
    \caption{Hierarchical taxonomy of Sparse Dictionary Learning research in Mechanistic Interpretability.}
    \label{fig:taxonomy}
\end{figure}

\newpage
\section{Notations}
\label{app:notation}

\begin{table}[H]
\centering
\small
\begin{tabular}{ll}
\toprule
\textbf{Notation} & \textbf{Description} \\
\midrule
\multicolumn{2}{l}{\textit{Distributions and Indexing}} \\
$\mathcal{D}$ & Distribution over inputs \\
$\mathcal{X}$ & Input space \\
$s$ & Sample drawn from $\mathcal{D}$ \\
$\mathbf{e}_d$ & Standard basis vector (1 in position $d$, 0 elsewhere) \\
\midrule
\multicolumn{2}{l}{\textit{Representations and Dimensions}} \\
$\mathbf{x}(s)$ & Ground-truth features for input $s \sim \mathcal{D}$ \\
$\mathbf{x}_p(s)$ & Input representation to SDL model (observed) \\
$\mathbf{x}_r(s)$ & Target/output representation for SDL model \\
$\mathbf{x}_q(s)$ & Latent activations in SDL bottleneck \\
$n$ & Number of ground-truth features (dimension of $\mathbf{x}$) \\
$n_p$ & Dimension of input representation $\mathbf{x}_p$ \\
$n_r$ & Dimension of target representation $\mathbf{x}_r$ \\
$n_q$ & Dimension of latent space $\mathbf{x}_q$ \\
\midrule
\multicolumn{2}{l}{\textit{SDL Architecture}} \\
$W_E$ & Encoder matrix ($n_q \times n_p$) \\
$W_D$ & Decoder matrix ($n_r \times n_q$) \\
$\sigma(\cdot)$ & Sparsity-inducing activation function \\
$\mathbf{w}_E^i$ & $i$-th row of encoder $W_E$ (encodes to neuron $i$) \\
$\mathbf{w}_D^i$ & $i$-th column of decoder $W_D$ (decodes from neuron $i$) \\
\midrule
\multicolumn{2}{l}{\textit{Linear Representation Hypothesis}} \\
$W_p$ & Feature matrix for $\mathbf{x}_p$ ($n_p \times n$) \\
$W_r$ & Feature matrix for $\mathbf{x}_r$ ($n_r \times n$) \\
$\mathbf{w}_p^d$ & $d$-th column of $W_p$ (feature direction for feature $d$) \\
$\mathbf{w}_r^d$ & $d$-th column of $W_r$ (feature direction for feature $d$) \\
$W_p^{\text{true}}$ & Ground-truth feature matrix (Linear Representation Bench) \\
$W_p^{\text{learned}}$ & Learned feature matrix ($W_E^\top$) \\
\midrule
\multicolumn{2}{l}{\textit{Sparsity and Interference}} \\
$S$ & Sparsity level: $\Pr(\mathbf{x}_i(s) = 0) \geq S$ \\
$M$ & Maximum interference: $\max_{i \neq j} \langle W_p[:, i], W_p[:, j] \rangle$ \\
$M_d$ & Weight for feature $d$: $\Pr(\mathbf{x}(s) = x_d(s)\mathbf{e}_d) \cdot \mathbb{E}[x_d(s)^2 | x_d(s) > 0]$ \\
\midrule
\multicolumn{2}{l}{\textit{Feature Anchoring}} \\
$k$ & Number of anchor features \\
$\tilde{\mathbf{w}}_p^{(i)}, \tilde{\mathbf{w}}_r^{(i)}$ & $i$-th anchor feature pair \\
$\lambda_{\text{anchor}}$ & Anchoring loss weight \\
$\mathcal{K}$ & Set of indices for selected anchor features \\
\midrule
\multicolumn{2}{l}{\textit{Loss Functions}} \\
$\mathcal{L}_{\text{SDL}}$ & Standard SDL reconstruction loss \\
$\tilde{\mathcal{L}}_{\text{SDL}}$ & Approximate SDL loss (extreme sparsity regime) \\
$\mathcal{L}_{\text{anchor}}$ & Feature anchoring penalty \\
$\mathcal{L}_{\text{SDL-FA}}$ & Anchored SDL objective: $\mathcal{L}_{\text{SDL}} + \lambda_{\text{anchor}}\mathcal{L}_{\text{anchor}}$ \\
\midrule
\multicolumn{2}{l}{\textit{Activation Patterns}} \\
$\mathcal{P} = (\mathcal{F}_1, \ldots, \mathcal{F}_N)$ & Activation pattern \\
$\mathcal{F}_i$ & Set of ground-truth features activating neuron $i$ \\
$\mathcal{A}(d)$ & Set of neurons activated by feature $d$ \\
$\Omega_{\mathcal{A}}$ & Activation pattern region (fixed activation structure) \\
\midrule
\multicolumn{2}{l}{\textit{Evaluation Metrics}} \\
$\mathbf{M}_{\text{GT}}$ & GT Recovery: fraction of features with $> \tau$ alignment \\
$\mathbf{M}_{\text{IP}}$ & Maximum Inner Product: mean best alignment per feature \\
$\tau$ & Threshold for GT Recovery (typically $\tau = 0.9$ or $0.95$) \\
\bottomrule
\end{tabular}
\end{table}

\newpage
\section{The Linear Representation Bench}
\label{app:bench}

\begin{figure}[H]
    \centering
    \includegraphics[width=\linewidth]{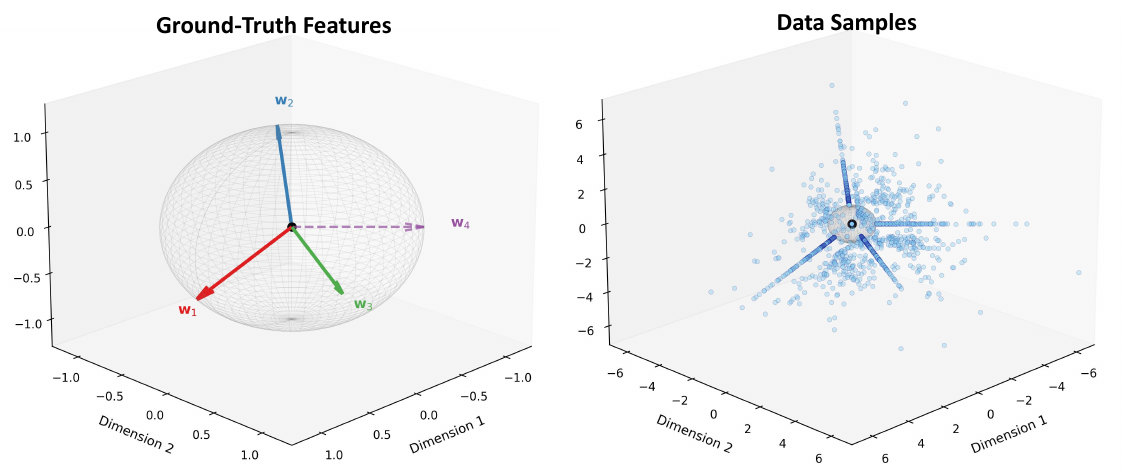}
    \caption{\textbf{Visualization of the Linear Representation Bench.} 
    The figure illustrates a $D=3$ dimensional representation space generated using $N=4$ feature directions ($\mathbf{W}_p$).}
    \label{fig:benchmark}
\end{figure}

The Linear Representation Bench is a synthetic benchmark designed to precisely instantiate the Representation Assumptions (Assumption~\ref{ass:lrh}) with fully accessible ground-truth features (See Figure~\ref{fig:benchmark} for a visualization of the benchmark). This enables rigorous evaluation of SDL methods under controlled conditions where feature recovery can be directly measured.

\subsection{Ground-Truth Feature Matrix Generation}

We construct the feature matrix $\mathbf{W}_p \in \mathbb{R}^{n_p \times n}$ to satisfy the representation assumptions (Assumption~\ref{ass:lrh}).

\paragraph{Initialization.} Initialize $\mathbf{W}_p$ with random Gaussian entries and normalize each column to unit $\ell_2$-norm:
\begin{equation}
    \mathbf{w}_p^d \leftarrow \frac{\mathbf{w}_p^d}{\|\mathbf{w}_p^d\|_2}, \quad \forall d \in [n]
\end{equation}

\paragraph{Interference Minimization.} We minimize pairwise interference via projected gradient descent. Define the soft-thresholded interference loss:
\begin{equation}
    \mathcal{L}_{\text{int}}(\mathbf{W}_p) = \sum_{i \neq j} \left[ \max\left(0, \langle \mathbf{w}_p^i, \mathbf{w}_p^j \rangle - (M - \epsilon) \right) \right]^2 + \lambda \sum_{i \neq j} \left[ \max\left(0, \langle \mathbf{w}_p^i, \mathbf{w}_p^j \rangle \right) \right]^2
\end{equation}
where $\epsilon > 0$ is a tolerance margin and $\lambda > 0$ is a small regularization weight encouraging negative interference.

At each iteration, we compute the gradient with respect to $\mathbf{W}_p$, perform a gradient step, and project back to the unit sphere:
\begin{equation}
    \mathbf{W}_p \leftarrow \mathbf{W}_p - \eta \nabla_{\mathbf{W}_p} \mathcal{L}_{\text{int}}, \quad
    \mathbf{w}_p^d \leftarrow \frac{\mathbf{w}_p^d}{\|\mathbf{w}_p^d\|_2}, \quad \forall d \in [n]
\end{equation}

This procedure continues until the maximum interference satisfies $\max_{i \neq j} \langle \mathbf{w}_p^i, \mathbf{w}_p^j \rangle \leq M$.

\subsection{Sparse Coefficient Generation}

For each sample $i \in [N]$, we generate sparse ground-truth features $\mathbf{x}^{(i)} \in \mathbb{R}^n_+$ as follows:

\paragraph{Sparsity Mask.} Each feature $d \in [n]$ is independently activated with probability $(1 - \mathcal{S})$:
\begin{equation}
    m_d^{(i)} \sim \text{Bernoulli}(1 - \mathcal{S})
\end{equation}

\paragraph{Feature Magnitudes.} Active features follow a shifted exponential distribution:
\begin{equation}
    x_d^{(i)} = m_d^{(i)} \cdot \left( c_{\min} + \text{Exp}(\beta) \right)
\end{equation}
where $c_{\min} \geq 0$ is a minimum activation threshold and $\beta > 0$ is the scale parameter.

This construction ensures:
\begin{itemize}
    \item \textbf{Non-negativity:} $x_d^{(i)} \geq 0$ for all $d, i$
    \item \textbf{Sparsity:} $\Pr(x_d^{(i)} = 0) = \mathcal{S}$ for all $d$
    \item \textbf{Independence:} Features $\{x_d^{(i)}\}_{d=1}^n$ are mutually independent
\end{itemize}

\subsection{Data Synthesis}

The observed representations are computed via linear combination:
\begin{equation}
    \mathbf{x}_p^{(i)} = \mathbf{W}_p \mathbf{x}^{(i)} = \sum_{d=1}^{n} x_d^{(i)} \mathbf{w}_p^d
\end{equation}

By construction, this dataset exactly satisfies Assumptions~\ref{ass:lrh}, enabling evaluation of SDL methods with complete knowledge of ground-truth features.

\subsection{Default Configuration}

Table~\ref{tab:bench_config} lists the default parameters used in our experiments.

\begin{table}[H]
\centering
\caption{Default configuration for the Linear Representation Bench.}
\label{tab:bench_config}
\begin{tabular}{lcc}
\toprule
\textbf{Parameter} & \textbf{Symbol} & \textbf{Default Value} \\
\midrule
Number of features & $n$ & 1000 \\
Representation dimension & $n_p$ & 768 \\
Superposition ratio & $n / n_p$ & 1.30$\times$ \\
Number of samples & $N$ & 100,000 \\
Sparsity level & $\mathcal{S}$ & 0.99 \\
Maximum interference & $M$ & 0.1 \\
\bottomrule
\end{tabular}
\end{table}

This controlled setting enables direct measurement of feature recovery metrics with ground-truth access, complementing evaluations on real neural network representations where true features are unknown.

\newpage
\section{Related Works}
\label{sec:related}

\subsection{Mechanistic Interpretability}

Interpretability is crucial for deploying AI in high-stakes domains such as medical diagnosis and financial modeling, where understanding model decisions is essential for safety and trust \citep{simon2024interplmdiscoveringinterpretablefeatures,abdulaal2024xrayworth15features,zhao2026rep2textdecodingtextsingle}. Traditional approaches include interpretability-by-design methods like Concept Bottleneck Models \citep{koh2020conceptbottleneckmodels} and decision trees \citep{mienye2024survey}. Concept Activation Vectors \citep{kim2018interpretabilityfeatureattributionquantitative} extend this by identifying human-defined concept directions in neural network representation spaces. Post-hoc explanation methods like GradCAM \citep{Selvaraju_2019} and SHAP \citep{lundberg2017unifiedapproachinterpretingmodel} provide local explanations without modifying the model architecture. Mechanistic interpretability \citep{sharkey2025openproblemsmechanisticinterpretability,bereska2024mechanisticinterpretabilityaisafety} aims to reverse-engineer neural networks by understanding their internal computational mechanisms. SAEs \citep{shu2025surveysparseautoencodersinterpreting} and related dictionary learning methods \citep{tang2025humanlikecontentanalysisgenerative,dunefsky2024transcodersinterpretablellmfeature} decompose neural activations into sparse, interpretable features. Circuit analysis \citep{olah2020zoom,olsson2022incontextlearninginductionheads} investigates how these features compose into algorithms.

\subsection{Sparse Dictionary Learning}

Sparse dictionary learning has a rich history predating its application to mechanistic interpretability. K-SVD \citep{aharon2006k} established foundational methods for learning overcomplete dictionaries, while theoretical work in compressed sensing \citep{donoho2006compressed} characterized recovery conditions, with \citet{spielman2012exactrecoverysparselyuseddictionaries} providing polynomial-time algorithms for exact reconstruction under sparsity assumptions. \citet{safran2018spuriouslocalminimacommon} demonstrated that spurious local minima are common even in simple two-layer ReLU networks, highlighting optimization challenges that persist in modern applications. Recent work has adapted these principles to mechanistic interpretability. SAEs \citep{cunningham2023sparseautoencodershighlyinterpretable} apply dictionary learning to language model activations, with LLMs widely utilized to analyze discovered features \citep{luo2024llm,luollm,tang2024demonstrationnotebookfindingsuited,zou2026fmlbenchbenchmarkingmachinelearning}. Various variants are further developed, including transcoders \citep{dunefsky2024transcodersinterpretablellmfeature}, crosscoders \citep{gao2024scalingevaluatingsparseautoencoders}, Matryoshka SAEs \citep{bussmann2025learningmultilevelfeaturesmatryoshka}, and hybrid approaches like Language-Grounded Sparse Encoders \citep{tang2025humanlikecontentanalysisgenerative}. These methods have found applications beyond language models, including protein structure analysis \citep{simon2024interplmdiscoveringinterpretablefeatures,gujral2025sparse}, medical imaging \citep{abdulaal2024xrayworth15features}, model evaluation \citep{tang2025doesmodelfailautomatic}, prompt engineering \citep{saini2026bridgingmechanisticinterpretabilityprompt}, board game analysis \citep{karvonen2024measuringprogressdictionarylearning}, and fMRI data analysis \citet{mao2025sparseautoencodersbridgedeep}.

\subsection{Biconvex Optimization}
Biconvex optimization studies problems where the objective is convex in each variable block when the other is fixed. \citet{gorski2007biconvex} provide a comprehensive survey establishing theoretical foundations and algorithmic approaches, while \citet{visweswaran1990global} develop the GOP algorithm providing global optimality guarantees via branch-and-bound. Matrix factorization problems exhibit similar bilinear structure; \citet{lee2009advances} introduce non-negative matrix factorization with multiplicative updates, and subsequent work establishes landscape properties, with \citet{ge2018matrixcompletionspuriouslocal} proving matrix completion has no spurious local minima and \citet{Sun_2016} providing convergence guarantees for alternating minimization. More broadly, \citet{Jain_2017} survey non-convex optimization in machine learning, while \citet{safran2018spuriouslocalminimacommon} demonstrate that spurious local minima are common in two-layer ReLU networks. Our work bridges these optimization-theoretic foundations with mechanistic interpretability by proving SDL exhibits biconvex structure, enabling the application of established algorithms and analysis techniques to this emerging field.
\newpage
\section{Limitations}
\label{sec:limitations}

While our theoretical framework provides valuable insights into SDL methods, several limitations warrant discussion:

\begin{itemize}
    \item \textbf{Assumption Violations.} Our analysis relies on the Representation Assumptions (Assumption~\ref{ass:lrh}), which may not hold perfectly in real-world neural networks. In particular, there exists features that are not one-dimentional linear \citep{engels2025languagemodelfeaturesonedimensionally} .
    
    \item \textbf{Extreme Sparsity.} Several key results, including Theorem~\ref{thm:loss-decomposition}, rely on the extreme sparsity assumption ($1- S < \frac{1}{n}$). The bounds may degrade substantially for moderate sparsity levels commonly observed in practice.
    
    \item \textbf{Feature Independence.} We assume mutual independence among ground-truth features, but real-world concepts often exhibit correlations that our analysis does not capture.
    
    \item \textbf{Convergence Guarantees.} We characterize global and partial minima of the optimization landscape but do not provide guarantees on whether gradient-based methods converge to global minima.
    
    \item \textbf{Anchor Availability.} Feature anchoring requires access to known semantic directions, which may not always be available. The quality and coverage of anchors significantly impact performance.
\end{itemize}

\section{Future Works}
\label{sec:future_works}

Our theoretical framework opens several avenues for future research:

\begin{itemize}
    \item \textbf{Global Optimization via GOP.} Having established that SDL is piecewise biconvex (Theorem~\ref{thm:biconvex}), applying Global Optimization for biconvex Problems (GOP)~\citep{gorski2007biconvex} could provide certificates of global optimality and systematically escape spurious partial minima.
    
    \item \textbf{Alternating Convex Search.} Alternating convex search (ACS)~\citep{gorski2007biconvex} directly exploits biconvex structure by alternately solving convex subproblems for $W_D$ and $W_E$. This may offer faster convergence and better solutions than standard gradient descent.
    
    \item \textbf{Convergence and Sample Complexity.} Building on our landscape characterization, future work should establish convergence rates for gradient descent and sample complexity bounds for feature recovery.
    
    \item \textbf{Moderate Sparsity Analysis.} Extending results beyond the $S \to 1$ regime to handle feature co-activation would broaden practical applicability.
    
    \item \textbf{Classical Dictionary Learning.} Adapting established algorithms like K-SVD~\citep{aharon2006k} and online dictionary learning~\citep{mairal2009online} to the SDL setting could yield improved optimization methods.
    
    \item \textbf{Automatic Anchor Discovery.} Developing methods to automatically discover high-quality anchors without external supervision would make feature anchoring more practical.
\end{itemize}
\newpage
\section{Ablation Studies}
\label{app:ablation}

Our theoretical analysis reveals that SDL optimization becomes increasingly underdetermined as the number of ground-truth features $n$ grows relative to the representation dimension $n_p$ (Theorem 3.4). To comprehensively validate our theoretical framework and assess the robustness of feature anchoring across different conditions, we conduct extensive ablation studies on the Linear Representation Bench, systematically varying superposition ratio, interference level, sparsity, and activation function parameters.

\paragraph{Effect of Superposition Ratio.} We first examine how feature anchoring performs under varying degrees of superposition by training ReLU SAEs with $n_q = 16,000$ latent dimensions while varying the number of ground-truth features $n \in \{800, 900, 1000, 1100, 1200, 1300, 1400\}$. The representation dimension is fixed at $n_p = n_r = 768$, yielding superposition ratios from $1.04\times$ to $1.82\times$. For feature anchoring, we randomly select $k = 100$ ground-truth features as anchors with $\lambda_{\text{anchor}} = 0.1$.

As shown in Table \ref{tab:anchor_ablation_ratio}, standard SAE training achieves 0\% GT Recovery across all feature counts, demonstrating the severity of the underdetermined optimization problem. With feature anchoring, GT Recovery reaches 100\% at $n = 800$ and remains high (98.89\%) at $n = 900$, confirming that anchoring a small subset of features (12.5\% and 11.1\% respectively) provides sufficient constraint to guide optimization toward the global minimum. As superposition increases, GT Recovery gradually decreases (85.50\% at $1.30\times$, 58.09\% at $1.43\times$, 30.58\% at $1.56\times$), aligning with our theoretical expectation that higher superposition expands the space of spurious solutions. Notably, feature anchoring consistently improves Maximum Inner Product across all settings, indicating better feature quality even when full recovery is not achieved.

\begin{table}[htbp]
    \centering
    \caption{Feature recovery under varying superposition ratios. We fix $n_p = n_r = 768$ and vary $n$ from 800 to 1400. Feature anchoring uses $k = 100$ randomly selected anchors with $\lambda_{\text{anchor}} = 0.1$.}
    \begin{tabular}{lccc}
    \toprule
         \textbf{Method} & \textbf{Num Features} & \textbf{GT Recovery ↑} & \textbf{Max Inner Product ↑} \\
    \midrule
         SAE & 800 & 0.00\% &  0.361\\
         \quad + Feature Anchoring &  800 &   \textbf{100.00\%} & \textbf{0.436} \\
    \midrule
         SAE & 900 & 0.00\% &  0.298\\
         \quad + Feature Anchoring &  900 & \textbf{98.89\%} & \textbf{0.345}\\
    \midrule
         SAE & 1000 & 0.00\% & 0.254\\
         \quad + Feature Anchoring &  1000 & \textbf{85.50\%} & \textbf{0.292}\\
    \midrule
         SAE & 1100 & 0.00\% &  0.223\\
         \quad + Feature Anchoring &  1100 & \textbf{58.09\%} & \textbf{0.258} \\
    \midrule
         SAE & 1200 & 0.00\% & 0.203\\
         \quad + Feature Anchoring &  1200 & \textbf{30.58\%} & \textbf{0.233}\\
    \midrule
         SAE & 1300 & 0.00\% & 0.192\\
         \quad + Feature Anchoring &  1300 & \textbf{15.23\%} & \textbf{0.216} \\
    \midrule
         SAE & 1400 & 0.00\% &  0.181\\
         \quad + Feature Anchoring &  1400 & \textbf{8.57\%} & \textbf{0.207}\\
    \bottomrule
    \end{tabular}
\label{tab:anchor_ablation_ratio}
\end{table}

\paragraph{Effect of Maximum Interference.} To examine how feature anchoring performs under different interference levels, we fix $n = 1000$ features in $n_p = n_r = 768$ dimensions and vary the maximum interference $M \in \{0.05, 0.1, 0.2, 0.5\}$ during ground-truth feature matrix generation. Recall that $M := \max_{i \neq j} \langle W_p[:, i], W_p[:, j] \rangle$ quantifies the maximum dot product between distinct feature directions, characterizing the degree of feature overlap in the representation space.

Table \ref{tab:anchor_ablation_interference} presents results across different interference levels. For TopK, BatchTopK, and Matryoshka SAEs, feature anchoring consistently improves GT Recovery across all interference values, with improvements ranging from 2-7 percentage points. For instance, at $M = 0.5$, BatchTopK SAE improves from 83.7\% to 89.6\% with anchoring. ReLU and JumpReLU SAEs, which struggle with feature recovery even with anchoring (0\% GT Recovery), still show improved Maximum Inner Product (e.g., ReLU at $M = 0.2$: 0.204 → 0.245), indicating better feature alignment. Interestingly, performance degrades slightly at very low interference ($M = 0.05$) compared to moderate interference ($M = 0.1, 0.2$), likely because extremely low interference creates near-orthogonal features that are easier to recover without additional constraints, making the anchoring less critical.

\begin{table}[htbp]
    \centering
    \caption{Feature recovery under varying maximum interference $M$. All experiments use $n = 1000$ features with $n_p = n_r = 768$ and $k = 100$ anchors.}
    \begin{tabular}{lcccc}
    \toprule
         \textbf{Method} & \textbf{Max Interference} & \textbf{GT Recovery ↑} & \textbf{Max Inner Product ↑} \\
    \midrule
        \multicolumn{4}{l}{\textit{Low Interference ($M = 0.05$)}} \\
         ReLU SAE & 0.05 & 0.0\% & 0.205 \\
         \quad + Feature Anchoring & 0.05 & 0.0\% & \textbf{0.246} \\
         JumpReLU SAE & 0.05 & 0.0\% & 0.239 \\
         \quad + Feature Anchoring & 0.05 & 0.0\% & \textbf{0.326} \\
         TopK SAE & 0.05 & 84.3\% & 0.982 \\
         \quad + Feature Anchoring & 0.05 & \textbf{87.3\%} & \textbf{0.985} \\
         BatchTopK SAE & 0.05 & 86.1\% & 0.984 \\
         \quad + Feature Anchoring & 0.05 & \textbf{87.5\%} & \textbf{0.986} \\
         Matryoshka SAE & 0.05 & 83.8\% & 0.982 \\
         \quad + Feature Anchoring & 0.05 & \textbf{85.8\%} & \textbf{0.985} \\
    \midrule
        \multicolumn{4}{l}{\textit{Moderate Interference ($M = 0.2$)}} \\
         ReLU SAE & 0.2 & 0.0\% & 0.204 \\
         \quad + Feature Anchoring & 0.2 & 0.0\% & \textbf{0.245} \\
         JumpReLU SAE & 0.2 & 0.0\% & 0.237 \\
         \quad + Feature Anchoring & 0.2 & 0.0\% & \textbf{0.333} \\
         TopK SAE & 0.2 & 85.2\% & 0.984 \\
         \quad + Feature Anchoring & 0.2 & \textbf{86.8\%} & \textbf{0.986} \\
         BatchTopK SAE & 0.2 & 84.7\% & 0.981 \\
         \quad + Feature Anchoring & 0.2 & \textbf{90.7\%} & \textbf{0.989} \\
         Matryoshka SAE & 0.2 & 84.7\% & 0.983 \\
         \quad + Feature Anchoring & 0.2 & \textbf{87.6\%} & \textbf{0.986} \\
    \midrule
        \multicolumn{4}{l}{\textit{High Interference ($M = 0.5$)}} \\
         ReLU SAE & 0.5 & 0.0\% & 0.203 \\
         \quad + Feature Anchoring & 0.5 & 0.0\% & \textbf{0.243} \\
         JumpReLU SAE & 0.5 & 0.0\% & 0.237 \\
         \quad + Feature Anchoring & 0.5 & 0.0\% & \textbf{0.325} \\
         TopK SAE & 0.5 & 84.2\% & 0.982 \\
         \quad + Feature Anchoring & 0.5 & \textbf{88.2\%} & \textbf{0.987} \\
         BatchTopK SAE & 0.5 & 83.7\% & 0.981 \\
         \quad + Feature Anchoring & 0.5 & \textbf{89.6\%} & \textbf{0.987} \\
         Matryoshka SAE & 0.5 & 83.9\% & 0.982 \\
         \quad + Feature Anchoring & 0.5 & \textbf{86.4\%} & \textbf{0.985} \\
    \bottomrule
    \end{tabular}
\label{tab:anchor_ablation_interference}
\end{table}

\paragraph{Effect of Feature Sparsity.} We investigate how feature sparsity $S$ affects SDL performance by varying $S \in \{0.005, 0.01, 0.05, 0.1\}$ where $S$ denotes the probability that each feature is inactive. Lower $S$ values correspond to denser activation patterns, while higher $S$ approaches the extreme sparsity regime analyzed in our theoretical results.

Table \ref{tab:anchor_ablation_sparsity} reveals several key insights. At very low sparsity ($S = 0.005$), where features co-activate frequently, feature anchoring provides the most dramatic improvements. JumpReLU SAE improves from 11.8\% to 74.8\% GT Recovery, and all TopK-family methods show 2-6 percentage point gains. This validates that anchoring is particularly valuable when the extreme sparsity assumption is violated and feature co-occurrence complicates optimization. At moderate sparsity ($S = 0.05$), performance drops significantly across all methods, as increased co-activation creates more complex interference patterns. At high sparsity ($S = 0.1$), approaching our theoretical regime, TopK-family methods achieve near-perfect recovery (98.8\%-99.7\%), with anchoring providing marginal improvements. Interestingly, Matryoshka SAE benefits least from anchoring at extreme sparsity, likely because its multi-scale architecture already provides sufficient constraints to recover features.

\begin{table}[htbp]
    \centering
    \caption{Feature recovery under varying sparsity levels $S$. All experiments use $n = 1000$ features with $n_p = n_r = 768$ and $k = 100$ anchors.}
    \begin{tabular}{lcccc}
    \toprule
         \textbf{Method} & \textbf{Sparsity} & \textbf{GT Recovery ↑} & \textbf{Max Inner Product ↑} \\
    \midrule
        \multicolumn{4}{l}{\textit{Very Low Sparsity ($S = 0.005$)}} \\
         ReLU SAE & 0.005 & 0.0\% & 0.231 \\
         \quad + Feature Anchoring & 0.005 & 0.0\% & \textbf{0.326} \\
         JumpReLU SAE & 0.005 & 11.8\% & 0.890 \\
         \quad + Feature Anchoring & 0.005 & \textbf{74.8\%} & \textbf{0.970} \\
         TopK SAE & 0.005 & 82.2\% & 0.978 \\
         \quad + Feature Anchoring & 0.005 & \textbf{84.3\%} & \textbf{0.981} \\
         BatchTopK SAE & 0.005 & 80.5\% & 0.976 \\
         \quad + Feature Anchoring & 0.005 & \textbf{83.2\%} & \textbf{0.980} \\
         Matryoshka SAE & 0.005 & 79.3\% & 0.974 \\
         \quad + Feature Anchoring & 0.005 & \textbf{86.1\%} & \textbf{0.983} \\
    \midrule
        \multicolumn{4}{l}{\textit{Moderate Sparsity ($S = 0.01$)}} \\
        ReLU SAE & 0.01 & 0.00\% & 0.205\\
        \quad + Feature Anchoring & 0.01 & 0.00\% & \textbf{0.246}\\
        JumpReLU SAE & 0.01 &  0.00\% & 0.237\\
        \quad + Feature Anchoring & 0.01 &  0.00\% & \textbf{0.333}\\
        TopK SAE & 0.01 & 84.90\% & 0.983 \\
        \quad + Feature Anchoring & 0.01 & \textbf{87.63\%} & \textbf{0.986}\\
        BatchTopK SAE & 0.01 & 84.80\% & 0.981\\
        \quad + Feature Anchoring & 0.01 & \textbf{89.38\%} &\textbf{ 0.988 }\\
        Matryoshka SAE & 0.01 & 83.70\%& 0.982\\
        \quad + Feature Anchoring & 0.01 & \textbf{87.32\%} & \textbf{ 0.985}\\
    \midrule
        \multicolumn{4}{l}{\textit{High Sparsity ($S = 0.1$)}} \\
         ReLU SAE & 0.1 & 0.0\% & 0.186 \\
         \quad + Feature Anchoring & 0.1 & 0.0\% & \textbf{0.191} \\
         JumpReLU SAE & 0.1 & 0.0\% & 0.191 \\
         \quad + Feature Anchoring & 0.1 & 0.0\% & \textbf{0.193} \\
         TopK SAE & 0.1 & 98.8\% & 0.981 \\
         \quad + Feature Anchoring & 0.1 & 98.4\% & 0.979 \\
         BatchTopK SAE & 0.1 & 95.6\% & 0.975 \\
         \quad + Feature Anchoring & 0.1 & \textbf{96.2\%} & \textbf{0.976} \\
         Matryoshka SAE & 0.1 & 99.7\% & 0.991 \\
         \quad + Feature Anchoring & 0.1 & 99.7\% & 0.991 \\
    \bottomrule
    \end{tabular}
\label{tab:anchor_ablation_sparsity}
\end{table}

\paragraph{Effect of TopK Sparsity Parameter.} Finally, we examine how the sparsity-inducing parameter $k$ in TopK activation affects feature recovery. We train TopK SAEs with $k \in \{32, 64, 128\}$, corresponding to average activation rates of 0.2\%, 0.4\%, and 0.8\% of the $n_q = 16,000$ latent dimensions.

As shown in Table \ref{tab:anchor_ablation_topk}, both standard and anchored TopK SAEs achieve perfect (100\%) GT Recovery at $k = 32$, where extreme sparsity closely matches our theoretical assumptions. At $k = 64$, GT Recovery remains high (93.4\%-93.9\%) with minimal difference between standard and anchored training, indicating that moderate sparsity provides sufficient constraint for feature recovery without additional anchoring. At $k = 128$, where activation density increases, feature anchoring becomes beneficial again, improving GT Recovery from 86.7\% to 88.0\%. The Maximum Inner Product remains consistently high ($>0.985$) across all settings, confirming that TopK activation is generally effective for feature recovery, with anchoring providing incremental benefits at higher $k$ values where the optimization becomes more challenging.

\begin{table}[htbp]
    \centering
    \caption{Feature recovery under varying TopK sparsity parameter $k$. All experiments use $n = 1000$ features with $n_p = n_r = 768$, $n_q = 16,000$, and $k = 100$ anchors.}
    \begin{tabular}{lcccc}
    \toprule
         \textbf{Method} & \textbf{TopK Parameter $k$} & \textbf{GT Recovery ↑} & \textbf{Max Inner Product ↑} \\
    \midrule
         TopK SAE & 32 & 100.0\% & 0.999 \\
         \quad + Feature Anchoring & 32 & \textbf{100.0\%} & \textbf{0.999} \\
    \midrule
         TopK SAE & 64 & 93.9\% & 0.992 \\
         \quad + Feature Anchoring & 64 & 93.4\% & 0.991 \\
    \midrule
         TopK SAE & 128 & 86.7\% & 0.986 \\
         \quad + Feature Anchoring & 128 & \textbf{88.0\%} & \textbf{0.987} \\
    \bottomrule
    \end{tabular}
\label{tab:anchor_ablation_topk}
\end{table}

\paragraph{Theoretical Interpretation.} These comprehensive ablation studies validate our theoretical framework across diverse conditions. The consistent failure of ReLU and JumpReLU SAEs without sufficient sparsity constraints corroborates Theorem 3.4: the underdetermined solution space admits infinitely many configurations achieving low reconstruction loss without recovering interpretable features. Feature anchoring addresses this by constraining encoder-decoder pairs, reducing degrees of freedom and steering optimization away from spurious partial minima (Theorem 3.7). The varying effectiveness across interference levels, sparsity regimes, and activation functions reflects the fundamental trade-off in our assumptions: more severe feature compression (higher $M$), denser activations (lower $S$), or insufficient activation sparsity (higher $k$) create more complex optimization landscapes where even anchored methods struggle to disentangle all features completely.

\newpage
\section{Qualitative Examples}
\label{sec:qualitative_examples}

\begin{figure}[H]
    \centering
    \includegraphics[width=0.8\linewidth]{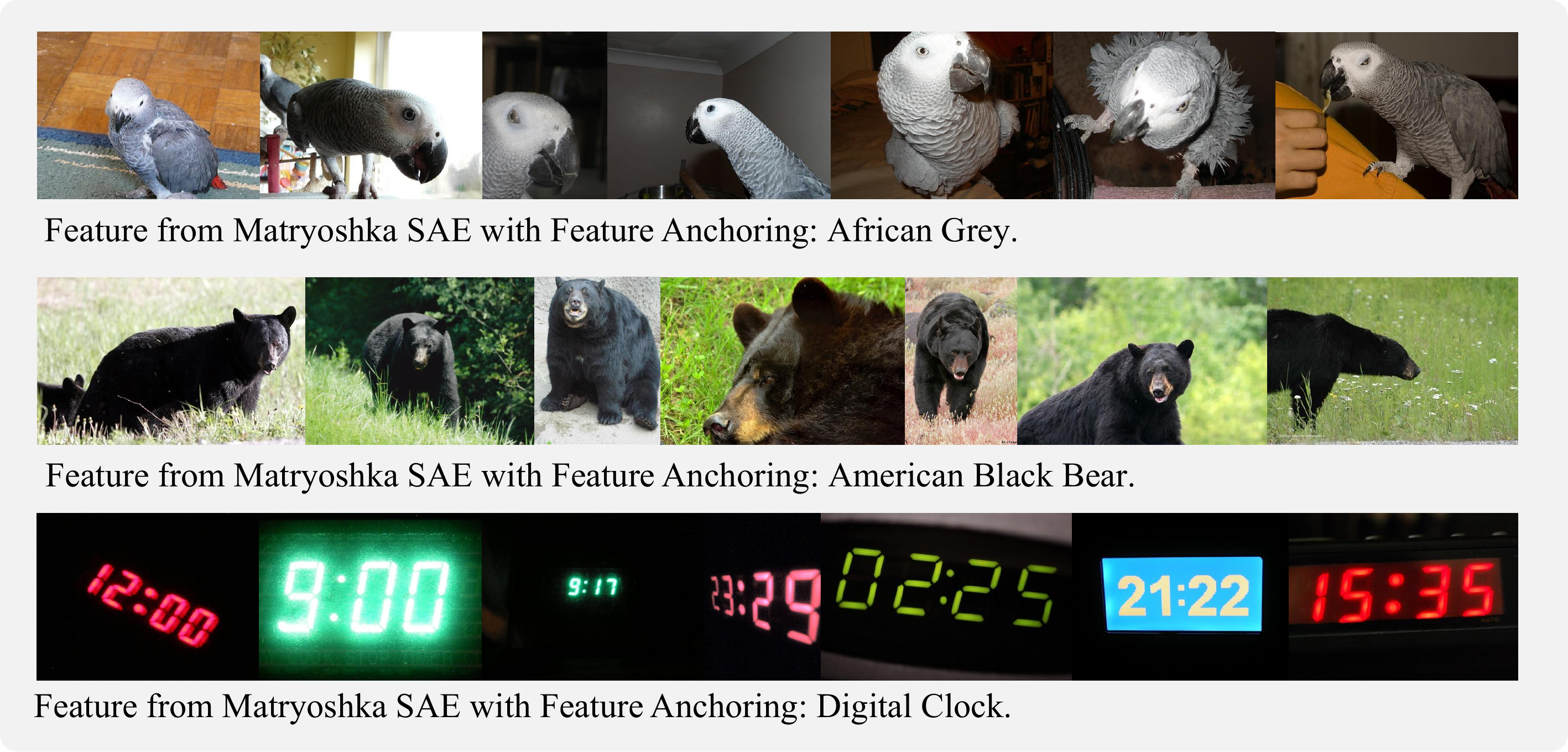}
    \caption{\textbf{Features learned with feature anchoring exhibit monosemanticity.} Each row shows the top-activating images for a single feature from Matryoshka SAE trained on CLIP with feature anchoring.}
    \label{fig:qualitative_fa}
\end{figure}

\begin{figure}[H]
    \centering
    \includegraphics[width=0.8\linewidth]{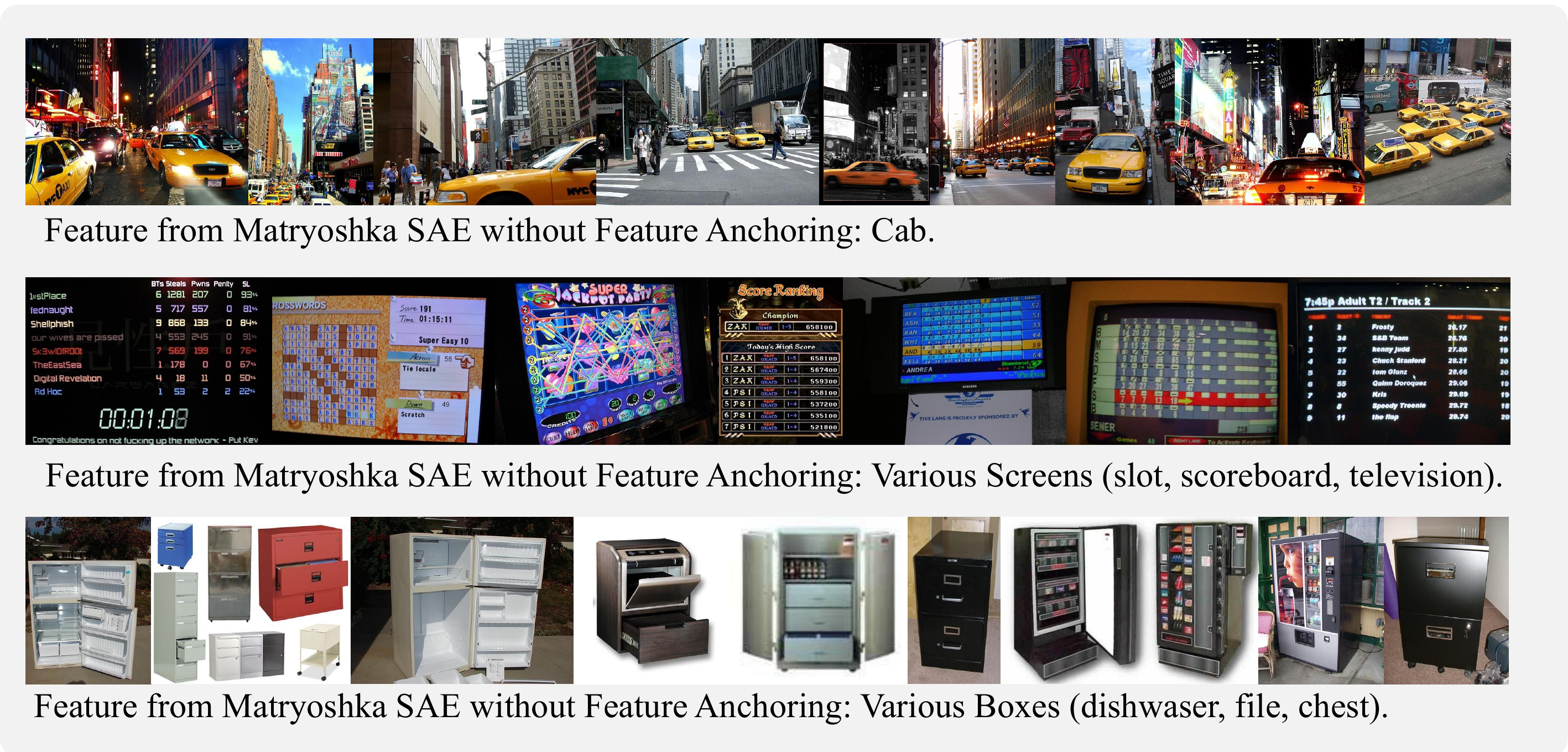}
    \caption{\textbf{Features learned without feature anchoring exhibit polysemanticity.} Each row shows the top-activating images for a single feature from Matryoshka SAE trained without feature anchoring.}
    \label{fig:qualitative_nofa}
\end{figure}

Beyond quantitative metrics, we provide qualitative evidence demonstrating how feature anchoring improves feature monosemanticity. Figure~\ref{fig:qualitative_fa} shows features learned by Matryoshka SAE with feature anchoring on CLIP, while Figure~\ref{fig:qualitative_nofa} shows features from the same architecture trained without feature anchoring.

\textbf{With Feature Anchoring.} As shown in Figure~\ref{fig:qualitative_fa}, the learned features exhibit clear monosemanticity: each feature responds to a single, well-defined visual concept. The ``African Grey'' feature activates exclusively on African Grey parrots, the ``American Black Bear'' feature captures only black bears, and the ``Digital Clock'' feature responds specifically to digital time displays. This monosemantic behavior aligns with our theoretical prediction that feature anchoring reduces the underdetermined nature of SDL optimization (Theorem~\ref{thm:zero-loss-conditions}).

\textbf{Without Feature Anchoring.} In contrast, Figure~\ref{fig:qualitative_nofa} illustrates the polysemanticity that emerges without anchoring. Features respond to multiple unrelated concepts: one activates on ``Various Boxes'' (dishwashers, file cabinets, chests), another on ``Various Screens'' (slot machines, scoreboards, televisions). While achieving low reconstruction loss, these features fail to capture semantically meaningful concepts---exactly the spurious partial minima characterized in Theorem~\ref{thm:local-existence}.

\vspace{-5mm}
\newpage
\section{Proofs of the Theorems}
\label{app:proofs}

\subsection{Proof of Theorem~\ref{thm:loss-decomposition}}
\label{app:loss-decomposition}
\begin{proof}
Let $C = \sup_{s \sim \mathcal{D}} \|\mathbf{x}_r(s) - W_D\sigma(W_E\mathbf{x}_p(s))\|_2^2 < \infty$
denote the uniform upper bound on the squared reconstruction error.
The loss decomposes over the number of active features:
\begin{align}
    \mathcal{L}_{\text{SDL}} 
    &= \sum_{m=0}^{n} \Pr(\|\mathbf{x}(s)\|_0 = m) 
       \cdot \mathbb{E}\bigl[\|\mathbf{x}_r(s) - W_D \sigma(W_E \mathbf{x}_p(s))\|^2 
       \mid \|\mathbf{x}(s)\|_0 = m\bigr].
\end{align}
The $m=0$ term contributes zero. For $m=1$, exactly one feature $d$ is active,
giving $\mathbf{x}_r(s) = x_d(s)w_r^d$ and $\mathbf{x}_p(s) = x_d(s)w_p^d$, so the
total contribution equals:
\begin{align}
    \sum_{d=1}^{n} M_d \left\| w_r^d - W_D \sigma(W_E w_p^d) \right\|^2 
    = \tilde{\mathcal{L}}_{\text{SDL}}(W_D, W_E),
\end{align}
where $M_d := \Pr(\mathbf{x}(s) = x_d(s)\mathbf{e}_d) \cdot \mathbb{E}[x_d(s)^2 \mid x_d(s) > 0]$.
For $m \geq 2$, since features are independently activated with probability $(1-S)$,
a union bound over all pairs $i \neq j$ gives:
\begin{equation}
    \Pr(\|\mathbf{x}(s)\|_0 \geq 2) 
    \leq \sum_{i,j}\Pr(x_i > 0,\, x_j > 0)
    \leq n^2(1-S)^2.
\end{equation}
Since the reconstruction error is bounded by $C$, combining all cases yields:
\begin{equation}
    \left|\mathcal{L}_{\text{SDL}} - \tilde{\mathcal{L}}_{\text{SDL}}\right| 
    \leq n^2 C (1-S)^2.
\end{equation}
\end{proof}

\subsection{Proof of Theorem~\ref{thm:biconvex}}
\label{app:biconvex}
\begin{proof}
\textbf{Preliminary observation.}
By definition of $\Omega_{\mathcal{P}}$ (Definition~\ref{def:activation-pattern}),
for any $W_E \in \Omega_{\mathcal{P}}$, the activation $z_d(W_E) := \sigma(W_E w_p^d)$
satisfies:
\begin{equation}
z_d(W_E)_i = \begin{cases}
\langle w_E^i, w_p^d \rangle & \text{if } i \in \mathcal{F}_d \\
0 & \text{otherwise,}
\end{cases}
\end{equation}
where $\mathcal{F}_d = \{i \in [n_q] : (\sigma(W_E w_p^d))_i > 0\}$.
Hence $z_d(W_E)$ is linear in $W_E$ within $\Omega_{\mathcal{P}}$.

\textbf{Per-feature decomposition.}
Define $f_d(W_D, W_E) := \|w_r^d - W_D z_d(W_E)\|^2$. Then:
\begin{equation}
\tilde{\mathcal{L}}_{\text{SDL}}(W_D, W_E) = \sum_{d=1}^n M_d f_d(W_D, W_E),
\quad M_d > 0.
\end{equation}
Since positive linear combinations preserve biconvexity, it suffices to show 
each $f_d$ is biconvex.

\textbf{Convexity in $W_D$.}
For fixed $W_E \in \Omega_{\mathcal{P}}$, $z_d$ is a fixed vector, so
$f_d(W_D) = \|w_r^d - W_D z_d\|^2$ is a squared affine function of $W_D$,
hence convex. Its Hessian $\nabla^2_{W_D} f_d = 2I_{n_r} \otimes (z_d z_d^\top) \succeq 0$
confirms this.

\textbf{Convexity in $W_E$.}
For fixed $W_D$, linearity of $z_d(W_E)$ in $W_E$ gives:
\begin{equation}
W_D z_d(W_E) = \sum_{i \in \mathcal{F}_d} \langle w_E^i, w_p^d \rangle w_D^i,
\end{equation}
which is affine in $W_E$. Hence $f_d(W_E) = \|w_r^d - W_D z_d(W_E)\|^2$ 
is a squared affine function of $W_E$, which is convex. For each row 
$w_E^i$ with $i \in \mathcal{F}_d$, the Hessian 
$\nabla^2_{w_E^i} f_d = 2\|w_D^i\|^2(w_p^d(w_p^d)^\top) \succeq 0$,
and rows $i \notin \mathcal{F}_d$ do not affect $f_d$.

\textbf{Conclusion.}
Each $f_d$ is biconvex, so $\tilde{\mathcal{L}}_{\text{SDL}}$ is biconvex 
over $\mathbb{R}^{n_r \times n_q} \times \Omega_{\mathcal{P}}$.
\end{proof}

\subsection{Proof of Theorem~\ref{thm:global-existence}}
\label{appendix:global-minimum}
\begin{proof}
We analyze $\tilde{\mathcal{L}}_{\text{SDL}}(W_D^*, W_E^*)$ by examining the 
reconstruction error for each feature $d \in [n]$ active in isolation.

The encoder output is $W_E^* w_p^d = W_p^\top w_p^d$ with $k$-th component
$(W_E^* w_p^d)_k = \langle w_p^k, w_p^d \rangle$.
By the unit-norm condition (Assumption~\ref{ass:lrh}), $(W_E^* w_p^d)_d = 1$.
Define the set of neurons activated by feature $d$:
\begin{equation}
    \mathcal{A}_d = \{k \in [n] : \sigma(W_E^* w_p^d)_k > 0\}.
\end{equation}
Since $\sigma(z)_i \in \{0, z_i\}$, we have $\sigma(W_E^* w_p^d)_k > 0$ if and 
only if $\langle w_p^k, w_p^d \rangle > 0$. Therefore $d \in \mathcal{A}_d$, and 
for $k \neq d$:
\begin{equation}
    \mathcal{A}_d \setminus \{d\} \subseteq \{k \in [n] : 0 < \langle w_p^k, w_p^d \rangle \leq M\}.
\end{equation}
The reconstruction is $W_D^*\sigma(W_E^* w_p^d) = \sum_{k \in \mathcal{A}_d} 
\langle w_p^k, w_p^d \rangle w_r^k$, so the reconstruction error for feature $d$ is:
\begin{align}
    w_r^d - W_D^* \sigma(W_E^* w_p^d) 
    &= w_r^d - \sum_{k \in \mathcal{A}_d} \langle w_p^k, w_p^d \rangle w_r^k 
     = -\sum_{k \in \mathcal{A}_d \setminus \{d\}} \langle w_p^k, w_p^d \rangle w_r^k.
\end{align}
Applying the triangle inequality and using $\|w_r^k\|_2 = 1$ and 
$\langle w_p^k, w_p^d \rangle \leq M$ for $k \in \mathcal{A}_d \setminus \{d\}$:
\begin{equation}
    \left\|w_r^d - W_D^* \sigma(W_E^* w_p^d)\right\|_2 
    \leq \sum_{k \in \mathcal{A}_d \setminus \{d\}} \langle w_p^k, w_p^d \rangle
    \leq M(K_d - 1),
\end{equation}
where $K_d = |\mathcal{A}_d| \leq n$. Squaring and summing over all features:
\begin{equation}
    \tilde{\mathcal{L}}_{\text{SDL}}(W_D^*, W_E^*) 
    \leq M^2 \sum_{d=1}^{n} M_d (K_d - 1)^2
    \leq n^2 M^2 \sum_{d=1}^{n} M_d,
\end{equation}
where the second inequality uses $(K_d - 1)^2 \leq (n-1)^2 \leq n^2$.
\end{proof}

\subsection{Proof of Theorem~\ref{thm:zero-loss-conditions}}
\label{appendix:zero-loss-conditions}

\begin{proof}
We prove both directions of the equivalence.

\textbf{Sufficient condition ($\Rightarrow$).} Suppose $w_r^d = W_D \sigma(W_E w_p^d)$ for all $d \in [n]$. By definition of the approximate loss:
\begin{align}
    \tilde{\mathcal{L}}_{\text{SDL}}(W_D, W_E) &= \sum_{d=1}^n M_d \| w_r^d - W_D \sigma(W_E w_p^d) \|^2 \\
    &= \sum_{d=1}^n M_d \| w_r^d - w_r^d \|^2 \\
    &= 0
\end{align}

\textbf{Necessary condition ($\Leftarrow$).} Suppose $\tilde{\mathcal{L}}_{\text{SDL}}(W_D, W_E) = 0$. Then:
\begin{equation}
    \sum_{d=1}^n M_d \| w_r^d - W_D \sigma(W_E w_p^d) \|^2 = 0
\end{equation}

Since $M_d > 0$ for all $d \in [n]$ (by definition, $M_d = \Pr(\mathbf{x}(s) = x_d(s)e_d) \cdot \mathbb{E}[x_d(s)^2 | x_d(s) > 0]$ where both factors are positive under Assumptions~\ref{ass:lrh}):
\begin{equation}
    \| w_r^d - W_D \sigma(W_E w_p^d) \|^2 = 0 \quad \forall d \in [n]
\end{equation}

Therefore:
\begin{equation}
    w_r^d = W_D \sigma(W_E w_p^d) \quad \forall d \in [n]
\end{equation}

\end{proof}

\subsection{Proof of Theorem~\ref{thm:local-existence}}
\label{appendix:local-minima}

\begin{proof}
We construct a configuration $(W_D^*, W_E^*)$ exhibiting the polysemanticity pattern and explicitly verify that both gradients vanish.

\textbf{Step 1: Given encoder $W_E^*$ from realizability.}

Since the activation pattern $\mathcal{P} = (\mathcal{F}_1, \ldots, \mathcal{F}_{n_q})$ is realizable and forms a partition of $[n]$, by Definition~\ref{def:activation-pattern}, there exists an encoder $W_E^* \in \mathbb{R}^{n_q \times n_p}$ such that:
\begin{equation}
    \forall i \in [n_q], \quad \mathcal{F}_i = \left\{ d \in [n] : \left(\sigma(W_E^* w_p^d)\right)_i > 0 \right\}
\end{equation}

Fix this $W_E^*$ and define the activation vectors:
\begin{equation}
    z_d := \sigma(W_E^* w_p^d) \in \mathbb{R}^{n_q}, \quad d = 1, \ldots, n
\end{equation}

Since $\mathcal{P}$ forms a partition, for each feature $d$ there exists a unique neuron $i(d)$ such that $d \in \mathcal{F}_{i(d)}$, and:
\begin{equation}
    z_d = \langle w_E^{*,i(d)}, w_p^d \rangle \cdot e_{i(d)}
\end{equation}

That is, only neuron $i(d)$ activates for feature $d$.

\textbf{Step 2: Construct optimal decoder $W_D^*$.}

For each neuron $i \in [n_q]$, we construct the decoder column $W_D^{*,i}$ by minimizing the loss over features activating that neuron.

\textit{Dead neurons:} If $\mathcal{F}_i = \emptyset$, set $W_D^{*,i} = 0$.

\textit{Active neurons:} If $\mathcal{F}_i \neq \emptyset$, solve:
\begin{equation}
    W_D^{*,i} = \arg\min_{W_D^i \in \mathbb{R}^{n_r}} \sum_{d \in \mathcal{F}_i} M_d \left\| w_r^d - W_D^i (z_d)_i \right\|^2
\end{equation}

This is a least squares problem with solution:
\begin{equation}\label{eq:optimal-decoder}
    W_D^{*,i} = \frac{\sum_{d \in \mathcal{F}_i} M_d (z_d)_i w_r^d}{\sum_{d \in \mathcal{F}_i} M_d (z_d)_i^2}
\end{equation}

\textbf{Step 3: Verify $\nabla_{W_D} \tilde{\mathcal{L}}_{\text{SDL}}(W_D^*, W_E^*) = 0$.}

The loss is $\tilde{\mathcal{L}}_{\text{SDL}}(W_D, W_E) = \sum_{d=1}^n M_d \|w_r^d - W_D z_d\|^2$. The gradient with respect to column $W_D^i$ is:
\begin{equation}
    \nabla_{W_D^i} \tilde{\mathcal{L}}_{\text{SDL}} = -2 \sum_{d=1}^{n} M_d (w_r^d - W_D z_d) (z_d)_i
\end{equation}

Since $\mathcal{P}$ is a partition, $(z_d)_i \neq 0$ only when $i = i(d)$, i.e., when $d \in \mathcal{F}_i$. Therefore:
\begin{equation}
    \nabla_{W_D^i} \tilde{\mathcal{L}}_{\text{SDL}}(W_D^*, W_E^*) = -2 \sum_{d \in \mathcal{F}_i} M_d (w_r^d - W_D^{*,i} (z_d)_i) (z_d)_i
\end{equation}

By construction (\ref{eq:optimal-decoder}), $W_D^{*,i}$ satisfies the normal equation:
\begin{equation}
    \sum_{d \in \mathcal{F}_i} M_d (z_d)_i (w_r^d - W_D^{*,i} (z_d)_i) = 0
\end{equation}

Therefore $\nabla_{W_D^i} \tilde{\mathcal{L}}_{\text{SDL}}(W_D^*, W_E^*) = 0$ for all $i \in [n_q]$.

\textbf{Step 4: Verify $\nabla_{W_E} \tilde{\mathcal{L}}_{\text{SDL}}(W_D^*, W_E^*) = 0$.}

Within the activation pattern region $\Omega_{\mathcal{A}}$, for feature $d \in \mathcal{F}_i$, the activation is:
\begin{equation}
    z_d = \langle w_E^{i}, w_p^d \rangle \cdot e_i
\end{equation}

The gradient with respect to encoder row $w_E^i$ is:
\begin{equation}
    \nabla_{w_E^i} \tilde{\mathcal{L}}_{\text{SDL}} = \sum_{d=1}^n M_d \nabla_{w_E^i} \left\|w_r^d - W_D z_d\right\|^2
\end{equation}

Only features $d \in \mathcal{F}_i$ have non-zero contribution. For such $d$:
\begin{equation}
    \frac{\partial z_d}{\partial w_E^i} = w_p^d \otimes e_i
\end{equation}

where $\otimes$ denotes outer product. Therefore:
\begin{align}
    \nabla_{w_E^i} \left\|w_r^d - W_D z_d\right\|^2 &= -2(w_r^d - W_D z_d)^\top W_D \frac{\partial z_d}{\partial w_E^i} \\
    &= -2(w_r^d - W_D^{*,i} \langle w_E^{*,i}, w_p^d \rangle)^\top W_D^{*,i} w_p^d
\end{align}

The total gradient is:
\begin{align}
    \nabla_{w_E^i} \tilde{\mathcal{L}}_{\text{SDL}} &= -2 \sum_{d \in \mathcal{F}_i} M_d (w_r^d - W_D^{*,i} \langle w_E^{*,i}, w_p^d \rangle)^\top W_D^{*,i} w_p^d \\
    &= -2 (W_D^{*,i})^\top \sum_{d \in \mathcal{F}_i} M_d \langle w_E^{*,i}, w_p^d \rangle (w_r^d - W_D^{*,i} \langle w_E^{*,i}, w_p^d \rangle)
\end{align}

From the normal equation in Step 3 with $(z_d)_i = \langle w_E^{*,i}, w_p^d \rangle$:
\begin{equation}
    \sum_{d \in \mathcal{F}_i} M_d \langle w_E^{*,i}, w_p^d \rangle (w_r^d - W_D^{*,i} \langle w_E^{*,i}, w_p^d \rangle) = 0
\end{equation}

Therefore:
\begin{equation}
    \nabla_{w_E^i} \tilde{\mathcal{L}}_{\text{SDL}}(W_D^*, W_E^*) = -2 (W_D^{*,i})^\top \cdot 0 = 0
\end{equation}

This holds for all active neurons. For dead neurons ($\mathcal{F}_i = \emptyset$), the gradient is automatically zero.

\textbf{Step 5: $(W_D^*, W_E^*)$ is a partial optimum.}

By Theorem~\ref{thm:biconvex}, $\tilde{\mathcal{L}}_{\text{SDL}}$ is convex in $W_D$ for fixed $W_E^*$, and convex in $W_E$ for fixed $W_D^*$ within $\Omega_{\mathcal{A}}$. 

Since we have verified:
\begin{itemize}
    \item $\nabla_{W_D} \tilde{\mathcal{L}}_{\text{SDL}}(W_D^*, W_E^*) = 0$ (Step 3)
    \item $\nabla_{W_E} \tilde{\mathcal{L}}_{\text{SDL}}(W_D^*, W_E^*) = 0$ within $\Omega_{\mathcal{A}}$ (Step 4)
\end{itemize}

By the first-order optimality condition for convex functions:
\begin{itemize}
    \item $W_D^*$ minimizes $\tilde{\mathcal{L}}_{\text{SDL}}(\cdot, W_E^*)$
    \item $W_E^*$ minimizes $\tilde{\mathcal{L}}_{\text{SDL}}(W_D^*, \cdot)$ over $\Omega_{\mathcal{A}}$
\end{itemize}

By standard results in biconvex optimization \citep{gorski2007biconvex}, $(W_D^*, W_E^*)$ is a partial optimum.

\textbf{Step 6: Non-zero loss.}

Since the pattern is polysemantic, there exists $i \in [n_q]$ with $|\mathcal{F}_i| \geq 2$. Choose distinct $d_1, d_2 \in \mathcal{F}_i$.

By Theorem~\ref{thm:zero-loss-conditions}, zero loss requires:
\begin{equation}
    w_r^{d_1} = W_D^* z_{d_1} = W_D^{*,i} (z_{d_1})_i, \quad w_r^{d_2} = W_D^* z_{d_2} = W_D^{*,i} (z_{d_2})_i
\end{equation}

If both equations held, then:
\begin{equation}
    \frac{w_r^{d_1}}{(z_{d_1})_i} = W_D^{*,i} = \frac{w_r^{d_2}}{(z_{d_2})_i}
\end{equation}

Since $(z_{d_1})_i = \langle w_E^{*,i}, w_p^{d_1} \rangle > 0$ and $(z_{d_2})_i = \langle w_E^{*,i}, w_p^{d_2} \rangle > 0$, this requires:
\begin{equation}
    w_r^{d_1} = \frac{(z_{d_1})_i}{(z_{d_2})_i} w_r^{d_2}
\end{equation}

This means $w_r^{d_1} \propto w_r^{d_2}$, contradicting Assumption~\ref{ass:lrh} (distinct features have linearly independent reconstruction vectors). Therefore:
\begin{equation}
    \tilde{\mathcal{L}}_{\text{SDL}}(W_D^*, W_E^*) = \sum_{d=1}^{n} M_d \|w_r^d - W_D^* z_d\|^2 > 0
\end{equation}

\textbf{Conclusion.} We have constructed $(W_D^*, W_E^*)$ exhibiting the polysemanticity pattern, explicitly verified both gradients vanish, and shown it achieves positive loss, completing the proof.
\end{proof}

\subsection{Proof of Theorem~\ref{thm:hierarchical-absorption}}
\label{appendix:hierarchical-absorption}

\begin{proof}
We construct a realizable pattern exhibiting feature absorption by scaling down the parent neuron's encoder until one sub-concept separates, then adding a dedicated neuron for the separated feature.

\textbf{Step 1: Initial configuration and feature geometry.}

Since $\mathcal{P} = (\mathcal{F}_1, \ldots, \mathcal{F}_M)$ is realizable, there exists an encoder $W_E \in \mathbb{R}^{M \times n_p}$ and threshold $c > 0$ such that:
\begin{equation}
    \forall i \in [M], \quad \mathcal{F}_i = \{d \in [n] : \langle w_E^i, w_p^d \rangle > c\}
\end{equation}

Fix any parent neuron $i^* \in [M]$ with sub-concepts $\mathcal{F}_{i^*} = \{d_{i^*,1}, \ldots, d_{i^*,k_{i^*}}\}$ where $k_{i^*} \geq 2$. 

For each sub-concept, compute its activation strength:
\begin{equation}
    a_j := \langle w_E^{i^*}, w_p^{d_{i^*,j}} \rangle > c, \quad \forall j \in [k_{i^*}]
\end{equation}

Since there are finitely many sub-concepts, define:
\begin{equation}
    a_{\min} := \min_{j \in [k_{i^*}]} a_j > c
\end{equation}

Let $j^* \in [k_{i^*}]$ be the index achieving this minimum: $a_{j^*} = a_{\min}$.

\textbf{Step 2: Scale down parent neuron to separate one sub-concept.}

We shrink the encoder row $w_E^{i^*}$ by multiplying with scaling factor $\lambda \in (0,1)$:
\begin{equation}
    \tilde{w}_E^{i^*} := \lambda \cdot w_E^{i^*}
\end{equation}

The new activation strengths become:
\begin{equation}
    \tilde{a}_j := \langle \tilde{w}_E^{i^*}, w_p^{d_{i^*,j}} \rangle = \lambda \cdot a_j
\end{equation}

Choose $\lambda$ such that:
\begin{equation}
    \lambda := \frac{c + a_{\min}}{2a_{\min}} \in \left(\frac{c}{a_{\min}}, 1\right)
\end{equation}

This gives:
\begin{itemize}
    \item For the minimum-activation feature: $\tilde{a}_{j^*} = \lambda a_{\min} = \frac{c + a_{\min}}{2} < c + \epsilon$ for any small $\epsilon > 0$
    \item For all other features: $\tilde{a}_j = \lambda a_j > \lambda a_{\min} = \frac{c + a_{\min}}{2}$
\end{itemize}

By choosing the threshold to be $c' = \frac{c + a_{\min}}{2}$, we have:
\begin{equation}
    \tilde{a}_{j^*} = c' \quad \text{(at boundary)}, \quad \tilde{a}_j > c' \text{ for all } j \neq j^*
\end{equation}

Slightly increasing $c'$ by infinitesimal $\epsilon > 0$ gives $c'' = c' + \epsilon$, ensuring:
\begin{equation}
    \tilde{a}_{j^*} < c'' < \tilde{a}_j \quad \forall j \neq j^*
\end{equation}

Therefore, with encoder row $\tilde{w}_E^{i^*}$ and threshold $c''$:
\begin{equation}
    \{d \in [n] : \langle \tilde{w}_E^{i^*}, w_p^d \rangle > c''\} = \mathcal{F}_{i^*} \setminus \{d_{i^*,j^*}\}
\end{equation}

\textbf{Step 3: Construct dedicated neuron for separated feature.}

We create a new encoder row parallel to the separated feature:
\begin{equation}
    w_E^{new} := \alpha \cdot w_p^{d_{i^*,j^*}}
\end{equation}
where $\alpha > 0$ is chosen sufficiently large.

By the unit-norm assumption ($\|w_p^{d_{i^*,j^*}}\| = 1$):
\begin{equation}
    \langle w_E^{new}, w_p^{d_{i^*,j^*}} \rangle = \alpha \|w_p^{d_{i^*,j^*}}\|^2 = \alpha
\end{equation}

For any other feature $d \neq d_{i^*,j^*}$:
\begin{equation}
    \langle w_E^{new}, w_p^d \rangle = \alpha \langle w_p^{d_{i^*,j^*}}, w_p^d \rangle \leq \alpha M
\end{equation}
where $M < 1$ is the maximum interference.

Choose $\alpha$ large enough such that:
\begin{equation}
    \alpha > c'', \quad \alpha M < c''
\end{equation}

This is possible by taking $\alpha = \frac{2c''}{1 + M}$, giving:
\begin{equation}
    \alpha = \frac{2c''}{1+M} > c'', \quad \alpha M = \frac{2c''M}{1+M} < c''
\end{equation}
where the first inequality uses $M < 1$ and the second uses $M < 1$.

Therefore:
\begin{itemize}
    \item $\langle w_E^{new}, w_p^{d_{i^*,j^*}} \rangle = \alpha > c''$ (separated feature activates new neuron)
    \item $\langle w_E^{new}, w_p^d \rangle \leq \alpha M < c''$ for all $d \neq d_{i^*,j^*}$ (no other features activate)
\end{itemize}

\textbf{Step 4: The new activation pattern.}

Construct the expanded encoder $W_E' \in \mathbb{R}^{(M+1) \times n_p}$:
\begin{equation}
    W_E' = \begin{bmatrix} 
        w_E^1 \\ 
        \vdots \\ 
        w_E^{i^*-1} \\
        \tilde{w}_E^{i^*} \\
        w_E^{i^*+1} \\
        \vdots \\
        w_E^M \\
        w_E^{new}
    \end{bmatrix}
\end{equation}

With threshold $c''$, the resulting activation pattern is:
\begin{equation}
    \mathcal{P}' = (\mathcal{F}_1, \ldots, \mathcal{F}_{i^*-1}, \mathcal{F}_{i^*} \setminus \{d_{i^*,j^*}\}, \mathcal{F}_{i^*+1}, \ldots, \mathcal{F}_M, \{d_{i^*,j^*}\})
\end{equation}

This pattern is realizable by construction.
\end{proof}

\end{document}